\documentclass[10pt,twocolumn,letterpaper]{article}

\usepackage{iccv}
\usepackage{times}
\usepackage{epsfig}
\usepackage{graphicx}
\usepackage{amsmath}
\usepackage{amssymb}
\usepackage{array}
\usepackage{pifont}
\usepackage{capt-of,etoolbox}
\usepackage{multirow}
\usepackage{cite}
\usepackage{hhline}
\usepackage{authblk}
\usepackage{bbding}
\usepackage{algorithm}
\usepackage{algorithmic}
\usepackage[utf8x]{inputenc} 

\usepackage[pagebackref=true,breaklinks=true,letterpaper=true,colorlinks,bookmarks=false]{hyperref}

\iccvfinalcopy % *** Uncomment this line for the final submission

\def\iccvPaperID{6341} % *** Enter the CVPR Paper ID here

\ificcvfinal\fi
\begin{document}

%%%%%%%%% TITLE
%\title{Accurate Facial Geometry Prediction: 3D Facial Alignment, Orientation Estimation, and 3D Face Models}
\title{Accurate 3D Facial Geometry Prediction by Multi-Task, Multi-Modal, and Multi-Representation Landmark Refinement Network}

\vspace{-15pt}
\author{Cho-Ying Wu \: Qiangeng Xu \: Ulrich Neumann\\
University of Southern California \\
{\tt\small \{choyingw, qiangenx, uneumann\}@usc.edu}
\vspace{-15pt}
}
\maketitle
%\thispagestyle{empty}

%%%%%%%%% ABSTRACT
\begin{abstract}
\vspace{-10pt}
This work focuses on complete 3D facial geometry prediction, including 3D facial alignment via 3D face modeling and face orientation estimation using the proposed multi-task, multi-modal, and multi-representation landmark refinement network (M$^3$-LRN). Our focus is on the important facial attributes, 3D landmarks, and we fully utilize their embedded information to guide 3D facial geometry learning. We first propose a multi-modal and multi-representation feature aggregation for landmark refinement. Next, we are the first to study 3DMM regression from sparse 3D landmarks and utilize multi-representation advantage to attain better geometry prediction. We attain the state of the art from extensive experiments on all tasks of learning 3D facial geometry. We closely validate contributions of each modality and representation. Our results are robust across cropped faces, underwater scenarios, and extreme poses. Specially we adopt only simple and widely used network operations in M$^3$-LRN and attain a near 20\% improvement on face orientation estimation over the current best performance. See our project page \href{https://choyingw.github.io/works/M3-LRN/index.html}{here}.
%\vspace{-30pt}
\end{abstract}

%%%%%%%%% BODY TEXT

%%% We also add acknowledgement here!!!!!
\section{Introduction}\footnotetext[1]{We thank Jingjing Zheng, Jim Thomas, and Cheng-Hao Kuo for their concrete comments and advice on this work.}
\label{intro}
Facial geometry prediction including 3D facial alignment, face orientation estimation, and 3D face modeling are fundamental tasks \cite{zhang2016joint, zollhofer2018state, tuan2018extreme, kim2018deep, wu2019mvf,deng2019accurate,tuan2017regressing,lv2017deep} and have applications on face recognition \cite{shi2006effective, juhong2017face}, tracking \cite{liu2017robust, deng2019menpo}, and compression \cite{wang2020one}. Recent works \cite{zhu2016face, zhu2019face, guo2020towards, tu20203d, feng2018joint} perform multi-task learning to predict 3D landmarks via 3D face modeling using 3DMM, which is a principal component analysis (PCA) describing 3D faces within shape and expression variational spaces. 3D faces can be constructed from estimated parameters and the associated PCA basis. Yet, face pose for these previous works is only a by-product without evaluation and discussion on relations with 3D landmarks and 3D face models. 

Further, \textit{3D facial landmarks} are important attributes to guide the learning of 3D facial geometry. Previous works \cite{zhu2016face, zhu2019face, tu20203d, guo2020towards, feng2018joint} only extract 3D landmarks directly from fitted 3D faces by 3DMM as alignment outputs to compute alignment losses. However, first, 3DMM is a parameterization method that constrains the flexibility of per-vertex deformation. Performance of direct landmark extraction from 3DMM- fitted faces as the output is limited. Second, their predicted landmarks are coarse without further refinement. Third, they do not consider the geometric information embedded in sparse 3D landmarks, which can provide approximate face shapes, and utilize the information to help 3DMM prediction.

\begin{figure}[bt!]
\begin{center}
\includegraphics[width=1.0\linewidth]{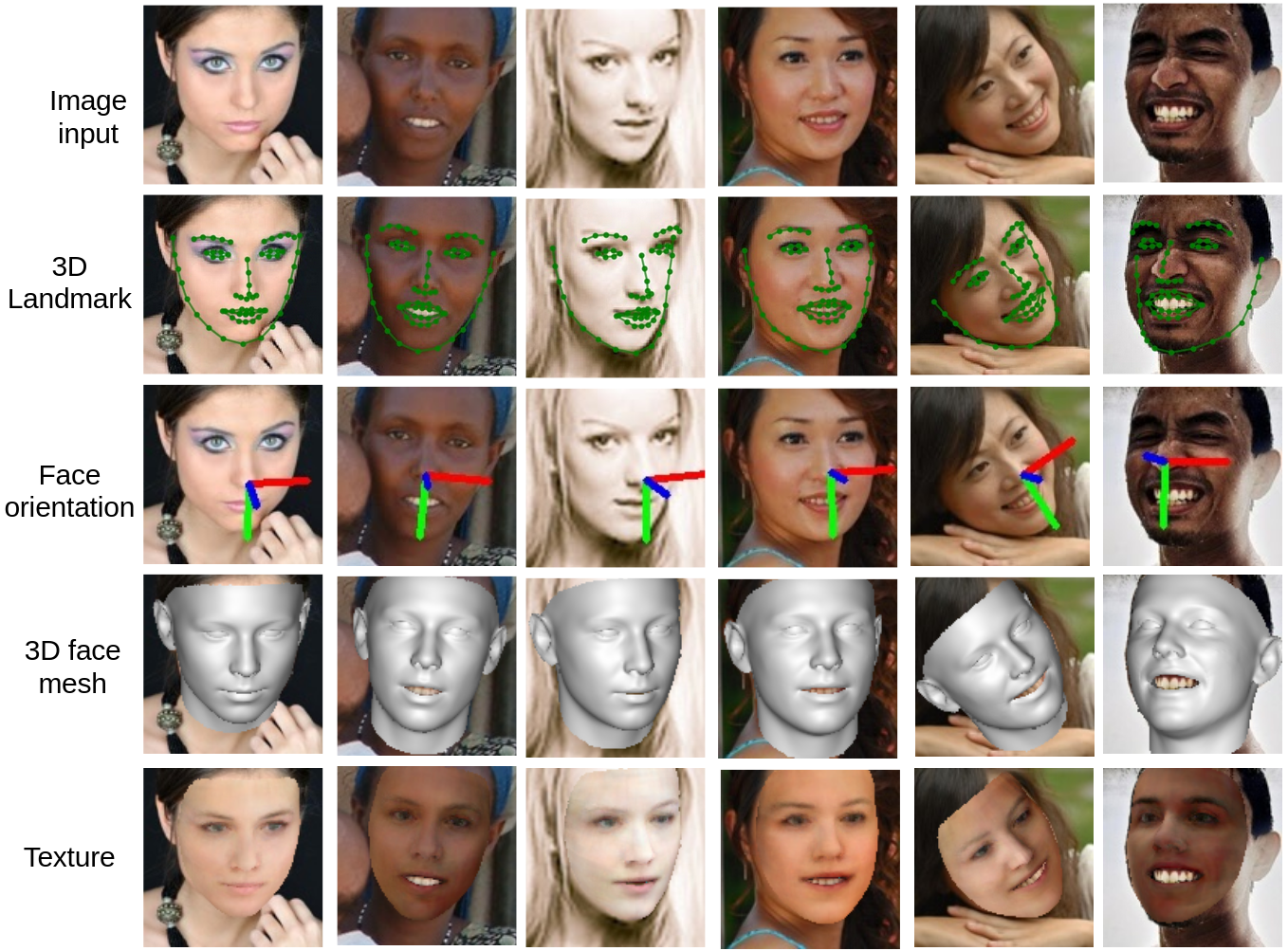}
\end{center}
\vspace{-13pt}
   \caption{\textbf{Results from our M$^3$-LRN with monocular image inputs.} Note that 3D landmarks can predict hidden face outlines in 3D rather than follow visible outlines on images.}
   \vspace{-11pt}
\label{teaser}
\end{figure}

On the other hand, recent works for face orientation estimation \cite{yang2019fsa, ruiz2018fine} are not robust since they only focus on the single task of Euler angle prediction from 2D images without exploiting 3D geometric information.

\begin{figure}[bt!]
\begin{center}
\includegraphics[width=1.0\linewidth]{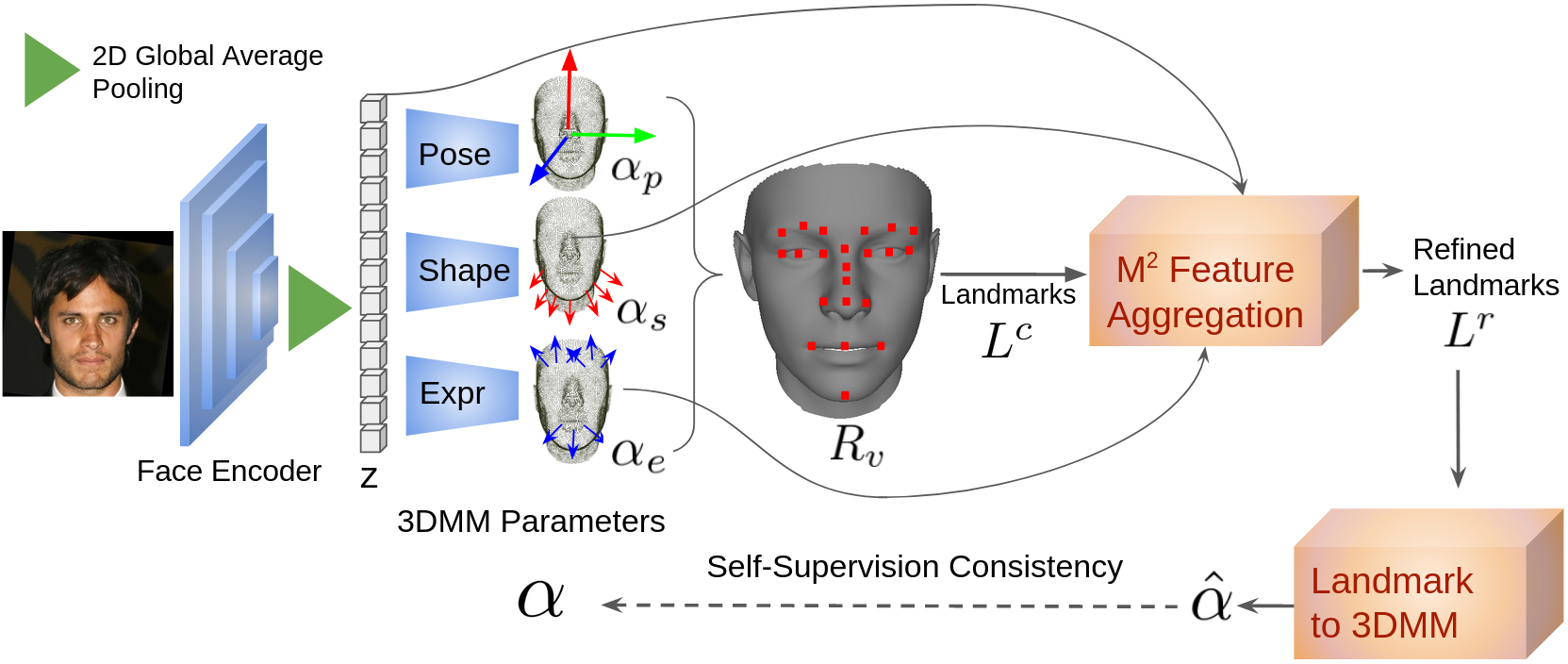}
\end{center}
\vspace{-12pt}
   \caption{\textbf{Pipeline of our method.} Backbone network learns to regress 3DMM parameters ($\alpha_p$,$\alpha_s$, and $\alpha_e$) and reconstruct 3D face meshes from monocular face images. M$^2$ Feature Aggregation (M$^2$FA) gathers underlying 3DMM semantics to further refine landmarks. The landmark-to-3DMM module regresses 3DMM from refined landmarks $L^r$ to learn embedded geometric information in 3D landmarks. A self-supervision consistency is applied to 3DMM regressed from different information sources. The red and blue arrows after shape and expression (expr) decoders show the main areas of deformation that each 3DMM semantics controls.}
   \vspace{-8pt}
\label{pipeline}
\end{figure}

To overcome the problems of the previous methods, in this work we propose \textbf{M$^3$-LRN}, a \textbf{multi-task} (3D alignment, 3D face modeling, and face orientation estimation), \textbf{multi-modal} (images, 3DMM modalities, and landmarks), and \textbf{multi-representation} (2D grid, 1D parameter, and 3D points) landmark refinement network that fully exploit landmarks to learn 3D facial geometry. Specially, we adopt a regress-then-refine approach to first regress 3DMM parameters and construct 3D meshes. After landmark extraction from meshes, we use the proposed multi-modal and multi-representation feature aggregation for landmark refinement. Next, we again \textit{estimate 3DMM parameters from holistic landmark features}. In this step, we leverage the embedded geometric information in sparse landmarks to help 3DMM prediction. The overall framework is in Fig. \ref{pipeline}.  

%a \textit{multi-task} (3D alignment, 3D face modeling, and face orientation estimation), \textit{multi-modal} (images, 3DMM modalities, and facial landmarks), and \textit{multi-representation} (2D grids, 1D parameters, and 3D point cloud) framework that focuses on complete 3D facial geometry prediction and evaluation. Specially, we \textit{go beyond the 3DMM process} with the proposed lightweight multi-modal and multi-representation refinement. Refined 3D landmarks are not restricted to lie on linear manifolds, which improves groundtruth fit while maintaining fast inference. The overall framework is shown in Fig. \ref{pipeline}. There are two stages, itemized 3DMM and landmark refinement with point geometry. 

%The itemized 3DMM (Sec.\ref{itemized3DMM}) disentangles underlying 3DMM modalities of pose, shape, and expression. Different from previous approaches that do not distinguish 3DMM parameter semantics \cite{zhu2019face, guo2020towards, zhu2016face, deng2019accurate}, we obtain improved controls for face topology deformation and pose estimation. Then after foundation face meshes are constructed, landmark refinement (Sec.\ref{landmark_PGS}) performs modality fusion to form a \textit{point-based multi-modal feature volume (MMFV)}, including landmark point features, image features, and shape and expression modalities. The MMFV is decoded to finer landmark structures. The refined landmarks are deformed from the foundation model to better fit groundtruth, and they are not constrained to lie on the low-rank 3DMM subspace.

We first review 3DMM in Sec.\ref{itemized3DMM} used in the previous work \cite{guo2020towards}. Our M$^3$-LRN estimates pose, shape, and expression parameters from a monocular face image through a backbone network. Meshes are then constructed as foundation models by 3DMM and landmarks are extracted by picked vertices. Next, for the refinement step (Sec.\ref{landmarkRef}), the proposed multi-modal and multi-representation feature aggregation (M$^2$FA) module performs modality fusion to build \text{multi-modal point feature} (MMPF), including landmark features, image features, and shape and expression modalities of 3DMM semantics. MMPF is then used to produce finer landmark structures. Advantage of MMPF is two-fold. First, landmark refinement based on only coarse landmarks is hard because the information is unitary. Joining different modalities and representations assists to refine and correct raw structures. For example, expression can better guide the refined landmarks to become more consistent to expressions that input faces appear. Second, the representation goes beyond parameterized 3DMM space to 3D points, which secure flexibility to learn per-landmark deformation for refinement compared with straight parameterization approach.

Furthermore, since 3D landmarks lying on eyes, mouth, and face outlines describe approximate facial geometry including shape, expression, and rotation. We assume 3D landmarks contain rough facial geometry. However 3D landmarks are different modality and representation from input images. Therefore, it is interesting to predict 3DMM parameters from landmarks to understand whether the multi-modal and multi-representation advantage help the performance. We predict 3DMM parameters from holistic landmark features by a landmark-to-3DMM regressor (Sec.\ref{pgs}). A self-supervised loss is introduced for 3DMM prediction of the same identity using information source. \textit{This is the first time 3DMM from sparse landmarks, referred to landmark geometry in this work, is studied}. We further explain the overall framework from a representation cycle view (Sec.\ref{cycle}).

Specially, our M$^3$-LRN contains only simple and widely used network operations. Our aim is to study how different modalities and representations contribute to the 3D facial geometry prediction on multitask learning. We quantitatively analyze performance gains introduced by each modality/representation with extensive experiments. We evaluate our M$^3$-LRN on facial alignment, face orientation estimation, and 3D face modeling on the standard datasets of each task. Our M$^3$-LRN attains the state of the art (SOTA) on all these tasks and achieve a near 20\% improvement on face orientation estimation over the current best work. Fig.\ref{teaser} demonstrates the ability of our M$^3$-LRN.

In summary, we present the following contributions:

1. We propose M$^3$-LRN, a multi-task (3D alignment, 3D face modeling, face orientation estimation), multi-modal (images, 3DMM modalities, landmarks), and multi-representation (2D grid, 1D parameter, and 3D points) framework using only simple and widely used operations to learn 3D facial geometry with SOTA performance.

2. We are the first to study landmark geometry and analyze the geometric information embedded in sparse 3D landmarks. We further introduce a novel self-supervision control for 3DMM estimation from images and landmarks.

%. We utilize the advantage of multi-modal and multi-representation learning for 3DMM prediction.

3. We conduct extensive and detailed evaluation of 3D facial geometry including facial alignment, face orientation estimation, and 3D face modeling to validate our SOTA performance on these tasks. Our error is near 20\% lower than the current best result for face orientation estimation.

%4. We also care about the application side of 3D facial alignment and propose a vertex selection approach and show the advantages of 3D over 2D, to advance the research of 3D approaches.

%For training, a novel data augmentation strategy is introduced (Sec.\ref{exp}), \textit{face swapping}, based on the motivation that face geometry is independent of face texture changes. We augment the dataset by swapping faces within a minibatch.

\section{Related Work}
\subsection{3D Facial Alignment via 3D Face Modeling}
\vspace{-5pt}
3D facial alignment aims at predicting 3D landmarks on images. In contrast, 2D approaches \cite{kazemi2014one, dong2018style, dong2019teacher,feng2018wing, wu2018look} usually regress direct landmark coordinates or heatmaps based on \textit{visible} facial parts. If input faces are self-occluded due to large face poses or faces are partially cropped out, their methods either only estimate landmarks along \textit{visible face outlines} (but not hidden outlines) or produce \textit{much larger error at invisible parts} and make their results unreliable. 

3D approaches \cite{zhu2016face, zhu2019face, feng2018joint, bulat2017far, guo2020towards, Jackson_2017_ICCV, shang2020self} predict 3D face models densely aligned with images. This way, occluded landmarks can also be registered. 3DDFA \cite{zhu2016face, zhu2019face} adopts BFM Face and use 3DMM fitting to reconstruct face meshes from monocular inputs. PRNet \cite{feng2018joint} predicts 2D UV-position maps that encode 3D points and uses BFM mesh connectivity to build face models. Compared with 3DDFA, PRNet might have higher mesh deformation ability, since its 3D points are not from 3DMM parameterization. However, it is harder to obtain a smooth and reliable mesh for PRNet. 2DASL \cite{tu20203d} based on 3DMM further adopts a differentiable renderer and a discriminator to produce high-quality 3D face models. 3DDFA-V2 \cite{guo2020towards} based on 3DDFA further introduces a meta-joint optimization strategy and a short video synthesis to attain the current best result. 3D faces and 3D landmarks from specified vertex indexing are outputs of these methods. However, their landmarks are raw and limited to the expressiveness of 3DMM basis, since 3DMM is a parameterization approach. In contrast, our method is a hybrid of 3DMM and point-cloud representations to secure more flexibility and better guide the facial geometry learning.

%However, these 3DMM fitting approaches construct face models lying on linear manifolds spanned by low-rank basis matrices. Reconstructed landmark structures are limited and dependent on expressiveness of the basis. In contrast, our method is a hybrid of 3DMM fitting and nonlinear landmark refinement to better fit groundtruth and attain SOTA.

\subsection{Face Orientation Estimation}
\vspace{-5pt}
Face orientation estimation has applications on human-robot interaction \cite{lemaignan2016real,wang2018human,palinko2016robot}. Euler angles (yaw, pitch, roll) are used to represent the orientation. Deep Head Pose \cite{mukherjee2015deep} uses networks to predict 2D landmarks and face orientation at the same time. HopeNet \cite{ruiz2018fine} uses bin-based angle regression to predict the Euler angles. FSA-Net \cite{yang2019fsa} constructs a fine-grained structure mapping for grouped features aggregation for better regression. These works focus on face orientation as a standalone task. On the other hand, although 3DMM-based 3D alignment approaches estimate rotation matrices, previous works only focus on evaluation and discussion of landmarks and 3D faces \cite{zhu2016face, zhu2019face, tu20203d, guo2020towards}. To gain an insight into full facial geometry, we benchmark both standalone orientation estimation methods and 3DMM-based approaches.

\section{Method}
Our method, illustrated in Fig. \ref{pipeline}, aims at precise and accurate 3D facial alignment, face orientation estimation, and 3D face modeling by \textit{fully utilizing facial landmarks} to better guide 3D facial geometry learning. The pipeline contains first-stage 3DMM regression with a backbone network, multi-modal/representation feature aggregation (M$^2$FA) for landmark refinement, and a landmark-to-3DMM regressor to utilize embedded geometric information in sparse landmarks.

%This pipeline contains an \textit{itemized 3DMM regression} for parameter regression and \textit{multi-modal landmark refinement with point geometry} using multi-modal and multi-representation information.

\subsection{3D Morphable Models (3DMM)}
\label{itemized3DMM}
3DMM reconstructs face meshes using PCA. Given a mean face $M \in \mathbb{R}^{3N_v}$ with $N_v$ 3D vertices, 3DMM deforms $M$ into a target face mesh by predicting the shape and expression variations. $U_{s}\in \mathbb{R}^{3N_v\times 40}$ is the basis for shape variation manifold that represents different identities, $U_{e}\in \mathbb{R}^{3N_v\times 10}$ is the basis for expression variation manifold, and $\alpha_{s}\in \mathbb{R}^{40}$ and $\alpha_{e}\in \mathbb{R}^{10}$ are the associated basis coefficients. The 3D face reconstruction can be formulated in Eq.\eqref{3DMM_basics}.
\begin{equation}
     R_{f} = Mat(M + U_{s}\alpha_{s} + U_{e}\alpha_{e}),
\label{3DMM_basics}
\end{equation}
where $R_{f}\in \mathbb{R}^{3\times N_v}$ represents a reconstructed frontal face model after the vector-to-matrix operation ($Mat$). To transform $R_{f}$ to input view, a 3x3 rotation matrix $P\in SO(3)$ and a translation vector $t\in \mathbb{R}^3$ are predicted to transform $R_f$ by Eq. (\ref{affine_trans}).
\begin{equation}
	R_v = PR_f+t,
\label{affine_trans}
\end{equation}
where $R_v\in \mathbb{R}^{3\times N_v}$ aligns with input view. $P$ and $t$ are included as 3DMM parameters in the most works \cite{wu2019mvf, zhu2016face, guo2020towards, guo2020towards}, and thus we use $\alpha_{p}\in \mathbb{R}^{12}$ instead. The whole 3DMM parameter $\alpha$ is 62-dim for pose, shape, and expression. Note that we follow the current best work 3DDFA-V2 \cite{guo2020towards} to predict 62-dim 3DMM parameters.

%We also adopt texture parameter regression $\alpha_{t}$ with associated basis to obtain colored face textures for better visualization.

%Previous 3DMM-based facial alignment \cite{tu20203d, zhu2016face, zhu2019face, zhu2019face, guo2020towards, wu2019mvf} predict $\alpha_{s}$ and $\alpha_{e}$ and use Eq. (\ref{3DMM_basics}) for 3D face modeling. However, they only treat 3DMM parameters as a bunch and use a single encoder-decoder network to regress $\alpha=\text{Dec}(z)$, with the bottleneck image feature $z$ and decoder Dec. They do not differentiate underlying parameter semantics and ignore the nature of 3DMM parameter regression is to obtain disentangled 3D face representation. 

We follow 3DDFA-V2 to adopt MobileNet-V2 as the backbone network to encode input images and use fully-connected (FC) layers as decoders for predicting 3DMM parameters from the bottleneck image feature $z$. We separate the decoder into several heads by 3DMM semantics which collaboratively predict the whole 62-dim parameters. The advantage of separate heads is that disentangling pose, shape, and expression controls secures better information flow. The illustration in Fig. \ref{pipeline} shows the encoder-decoder structure. The decoding is formulated as $\alpha_{m}=\text{Dec}_{m}(z)$, $m\in\{p, s, e\}$, showing pose, shape and expression. With groundtruth notation $^*$ hereafter, regression loss is shown as follows.
\vspace{-7pt}
\begin{equation}
     \mathbb{L}_{\text{3DMM}} = \sum_{m} \|\alpha_{m}-\alpha^*_{m}\|^2.
\label{loss_S1_3dmm}
\vspace{-7pt}
\end{equation}
After the regression, face meshes are constructed by Eq.\ref{3DMM_basics}-\ref{affine_trans} using popular BFM Face \cite{paysan20093d} including 53K vertices as \textit{foundation face models}. Then, $N_l$-point 3D landmarks $L^c \in \mathbb{R}^{3 \times N_l}$ are extracted from 3D vertices using specified indices.

%We design itemized 3DMM decoders to separate the decodings by the semantics of $\alpha$. Pose regression controls the view projection and does not deform the mesh topology. In contrast, shape and expression control the mesh deformation of the mean face. Therefore, we adopt a pose decoder and a topology deformation decoder consisting of 3 fully-connected (FC) layers and shape and expression decoders. The FC layers before the shape and expression decoders further learn features shared across them for better adaptation to the topology deformation. The structure is shown in Fig. \ref{pipeline}. The decodings are formulated as $\alpha_{m}=\text{Dec}_{m}(z)$, $m\in\{p, s, e\}$, showing pose, shape and expression. With groundtruth notation $^*$ hereafter, regression loss is shown as follows.

\subsection{Multi-Modal/Representation Refinement}
\label{landmarkRef}
Previous studies \cite{guo2020towards, zhu2019face, zhu2016face} directly use extracted landmarks from foundation face models to compute the alignment loss for learning 3D facial geometry. However, these extracted landmarks are raw and constrained to the 3DMM parameterization ability, which lead to the performance bottleneck. Therefore, we adopt a hybrid representation approach of 3DMM regression and a point-based landmark refinement to gain higher landmark deformation ability and fit groundtruth landmarks better. 

Landmarks can be seen as a sequence of 3D points. Weight-sharing multi-layer perceptrons (MLPs) are commonly used for extracting features from structured points. A common framework \cite{qi2017pointnet, qi2017pointnet++} uses an MLP-encoder to extract high-dimensional point features, and then use another MLP-decoder to regress per-point attributes from the high dimensional embedding. Our refinement module takes structured sparse landmarks $L^c$ as inputs, and use MLPs with batch normalization and ReLU to extract high dimensional point features. At the bottleneck, global point max-pooling is applied to obtain global point features. We then regress refined 3D landmark from the point features.

\begin{figure}[bt!]
    \centering
    \includegraphics[width=1.0\linewidth]{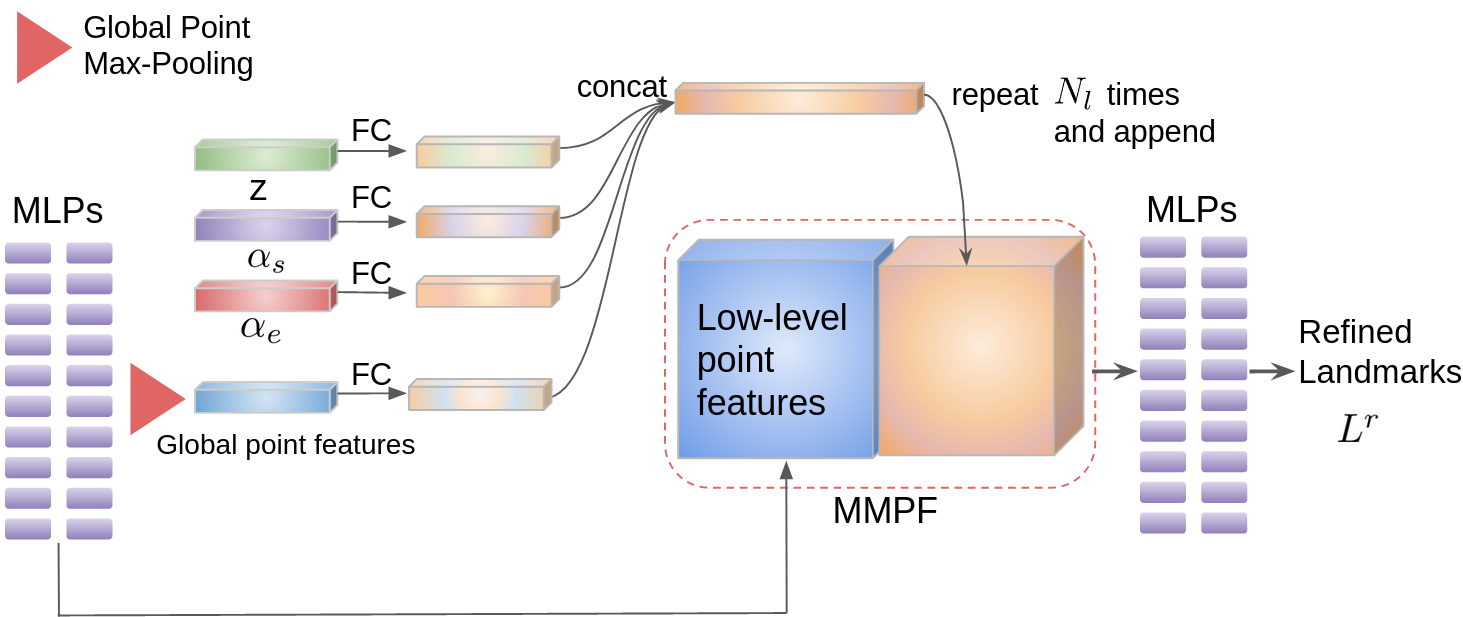}
    \vspace{-5pt}
    \caption{\textbf{Structure of M$^2$FA.} The input is $L^c$ from the foundation face model. The left MLPs extract global point features and fuse the global features with other modalities, including images features, shape, and expression parameters. The concatenated multi-modal features are appended to the low-level features to create MMPF, which is used to regress the refined landmarks.}
    \vspace{-10pt}
    \label{mmfv}
\end{figure}

In addition to directly using $L^c$ alone for the refinement, the input image and 3DMM modalities can provide information from the image, shape, and expression domains. Examples are shape contains information of thinner/thicker faces and expression contains information for eyebrow or mouth movements. Therefore, these pieces of information can help regress finer landmark structures. Specifically, Our M$^2$FA proposes to fuse information of the image, using its bottleneck feature $z$ after global average pooling, and shape and expression 3DMM parameters. These features and parameters are global information without spatial dimensions, and FC layers are applied to these modalities for adaption. 
Later we concatenate them into a multi-modal feature vector and then repeat this vector $N_l$ (number of landmarks) times. We last append the repeated features to the low-level landmark point features to create multi-modal point feature (MMPF). 
The overall design of M$^2$FA is shown in Fig. \ref{mmfv}. MMPF is later fed into an MLP-decoder to regress refined 3D landmarks. Skip connection is used from the coarser to refined landmarks to facilitate training. 

We use groundtruth landmarks to guide the training. The alignment loss function is formulated as follows.
\begin{equation}
     \mathbb{L}_{lmk} = \sum_{n} \mathbb{L}_{smL1}(L^r_n-L_n^*), n\in[1,N_l],
\label{loss_S1_lmk}
\vspace{-5pt}
\end{equation}
where $N_l$ is number of landmarks, $^*$ denotes groundtruth, and $\mathbb{L}_{smL1}$ is smooth L1 loss.

\subsection{Landmark Geometry Support (LGS)} 
\label{pgs}
Previous works only regress 3DMM parameters from images \cite{zhu2016face, zhu2019face,guo2020towards,tu20203d, deng2019accurate,wu2019mvf,Jackson_2017_ICCV}, where only single representation is used for 3DMM regression. However, facial landmarks are sparse keypoints lying at eyes, nose, mouth, and face outlines, which are principal areas that $\alpha_s$ and $\alpha_e$ control. We assume that approximate facial geometry is embedded in sparse landmarks. Thus, we further build a landmark-to-3DMM regressor to extract 3DMM parameters from the refined landmarks $L^r$ using its holistic landmark features. To our knowledge, we are the first to study geometric information embedded in sparse landmarks.

%We think that 3D landmarks contain auxiliary geometric information to describe the face structure with different modality and representation from source images. We predict 3DMM parameters from holistic landmark structure as support by a landmark-to-3DMM (LMK-3DMM) module (Sec.\ref{pgs}).

%Therefore 3DMM modalities of pose, shape, and expression are also embedded in the sparse landmark structure. From this observation, we further regress 3DMM parameters from the refined landmarks $L^r$. To our knowledge, we are the first to use deep networks to exploit landmark structures to regress 3DMM parameters.

%To our knowledge, we are the first to use deep networks to exploit sparse landmark structures to regress 3DMM parameters. Although landmark points are sparse and conceptually might not contain enough information for facial geometry estimation, sparse landmarks are extracted from dense mesh vertices, which is derived from 3DMM parameters at the first regression stage. More importantly, landmarks are facial keypoints lying at eyes, nose, mouth, and face outlines, which are principal areas $\alpha_{\text{sTDD}}$ and $\alpha_{\text{eTDD}}$ control. Therefore 3D attributes of pose, shape, and expression are also embedded in sparse landmark structure. 

%Pose $\alpha_{lp}$, shape $\alpha_{ls}$, and expression $\alpha_{le}$ are then produced from refined sparse landmarks $L^r$.

The landmark-to-3DMM regressor also contains an MLP-encoder to extract high dimensional point features and use a global point max-pooling to obtain holistic landmark features. Later separate FC layers as converters transform the holistic landmark features to 3DMM parameters to get $\hat{\alpha}$, including pose, shape, and expression modalities. We refer $\hat{\alpha}$ to \textit{landmark geometry}, since the facial geometry of 3DMM parameters is regressed from sparse landmarks. The supervised regression loss for landmark geometry is calculated as 
\begin{equation}
     \mathbb{L}_{\text{3DMM}_{lmk}} = \sum_{m} \|\hat{\alpha}_{m}-\alpha^*_m\|^2,
\label{loss_S2_3dmm}
\vspace{-7pt}
\end{equation}
where $m$ contains pose, shape, and expression.

Furthermore, since $\hat{\alpha}$ and $\alpha$ from the first stage describe the same identity, they should be numerically similar. We further propose a novel self-supervision control for 3DMM from images and landmarks as follows.
%and sparse landmarks guide each other for better training schema. 
\begin{equation}
\small
     \mathbb{L}_{g} = \sum_{m} \|\alpha_{m}-\hat{\alpha}_{m}\|^2,
\label{loss_S2_corr}
\vspace{-7pt}
\end{equation}
where $m \in \{p, s, e\}$. $\mathbb{L}_g$ improves information flow and lets facial geometry regressed from images obtain \textit{support} from landmarks.

The advantage of LGS is that since images and sparse landmarks are different data representations (2D grid and 3D points) using different network architectures and operations, more descriptive and richer features can be extracted and aggregated using a multi-representation framework. Although conceptually sparse landmarks provide rough face outlines, our experiments show that LGS further contributes to the final performance gain due to the multi-representation advantage. The total loss is combined as
%since our framework aggregates expressive features from multi-representations that fully utilizes the 
%we show that supportive information 
%PGS could further provides Therefore, our framework is a multi-modal and multi-representation framework for expressive feature extraction. 
\begin{equation}
\small
     \mathbb{L}_{total} = \lambda_1\mathbb{L}_{\text{3DMM}}+\lambda_2\mathbb{L}_{lmk}+\lambda_3\mathbb{L}_{\text{3DMM}_{lmk}}+\lambda_4\mathbb{L}_{g},
\label{loss_total}
\end{equation}
where $\lambda$ terms are loss weights. 

Compared with a simple baseline using only the first-stage 3DMM regression, our proposed refinement (M$^2$FA) and landmark geometry (LGS) only bring about 5\% more time in average for a single feed-forward pass, since landmarks are sparse ($N_l$=68 points for 300W-LP \cite{zhu2016face} as the training dataset) and weight-sharing MLPs are lightweight.

\subsection{Representation Cycle}
\label{cycle}
\begin{figure}[tb!]
    \centering
    \includegraphics[width=0.95\linewidth]{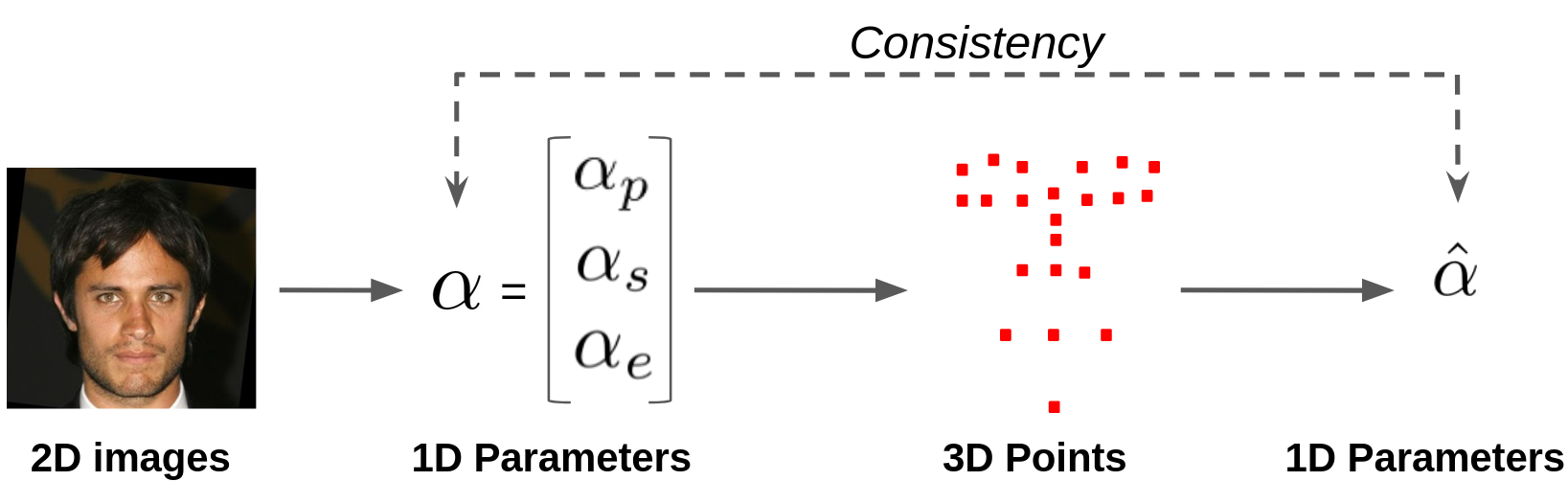}
    \caption{\textbf{Illustration of representation cycle.} }
    \label{representation_cycle}
    \vspace{-11pt}
\end{figure}

Cycle Consistency has been studied for domain adaption \cite{hoffman2018cycada} and knowledge transfer \cite{zhu2017unpaired}. It is widely used for data on different domains with the same representation, such as 2D images/label maps. Our M$^3$-LRN adopts cycle consistency on representations. Starting from 2D images, face encoder and separate decoders regress 1D parameters. After reconstructing 3D meshes from parameters and extracting 3D landmarks, the representation switches to 3D points. During the LGS step, the representation again switches from 3D points to 1D parameters. Therefore it forms a representation cycle (Fig. \ref{representation_cycle}) and we adopt the consistency to facilitate the training.

%the representation cycle consistency to facilitate the training.

%The strategy of our M$^3$-LRN is to aggregate multi-modal and multi-representation features from  multi-modal 

Overall, our network structure focuses on the multi-modal and multi-representation advantage for fully utilizing landmark information. We choose simple and widely used network operations to show that without special network designs, landmarks could better guide the 3D facial geometry learning. We validate our modality and representation choices through following studies and experiments. Network details are described in the supplementary.

% In literature sparse data are of research interests to assist learning from dense data sources such as images \cite{ma2018invertibility, zhong2019deep}.

\section{Experiments}
\label{exp}
Evaluation is conducted on the three focused tasks: facial alignment, face orientation estimation, and 3D face models. One of our main contribution is conducting extensive studies and exhibiting detailed performance gain breakdown for each module we propose. Our purpose is to analyze how each modality/representation contributes to the 3D facial geometry prediction using simple and widely used network operations such as MLPs and CNNs.

\textbf{Procedure.}
300W-LP\cite{zhu2016face} is a standard and widely used \textit{training} dataset on 3D face tasks. It collects in-the-wild face images and conducts face profiling \cite{zhu2016face} for producing 3DMM parameters with BFM Face \cite{paysan20093d} and FaceWarehouse texture \cite{cao2013facewarehouse}. The dataset also performs out-of-plane face rotation for augmentation, attaining more than 60K face images and fitted 3D models.

During training, we use a learning rate of 0.08, batch size of 1024, and momentum of 0.9 of the SGD \cite{zinkevich2010parallelized} optimizer. We train our network with 80 epochs, and the learning rate decays to $\frac{1}{10}$ and $\frac{1}{100}$ of the initial rate after 48 and 64 epochs. We use random color jittering and random flip with 0.5 of the probability. We also use a face swapping data augmentation for training on 300W-LP. The motivation is that groundtruth textures synthesized by 3DMM fitting are semantically aligned with mesh topology in this dataset, i.e., the $i$-th index for all fitted morphable models' vertex and texture always controls the same semantic point like nose tip or canthus. In this way, facial geometry is independent of texture changes. We randomly shuffle face textures within a minibatch of the training set and render 3D faces back onto images as the inputs. Groundtruth 3DMM parameters and landmarks are intact. We train on 4x GTX1080Ti GPUs, and the training takes 8 hours. %We leave the detailed network architectures to the supplementary.

At test time, the refined landmarks $L^r$, fitted 3D face $R_v$ with its orientation from $\alpha_p$ are the outputs for evaluation. The processing of landmark geometry is saved at test time since its information is auxiliary, and we leave the discussion and evaluation of landmark geometry in the supplementary. Our inference attains an average about 3000fps for the 3D landmark prediction and about 2600fps for the dense 3D face prediction on a single GPU with the MobileNet backbone. This satisfies the real-world applications for fast inference. 

Beside facial geometry, we also study texture synthesis in the supplementary for more realistic 3D faces and compare with textures synthesized by 3DMM texture fitting.

%, i.e., face textures would not affect groundtruth 3D geometry

%\textbf{Face Swapping.} We also propose a new data augmentation technique, face swapping. The motivation is that groundtruth textures in 300W-LP synthesized by 3DMM fitting are semantically aligned with mesh topology, i.e., the $i$-th index for all fitted morphable models' vertex and texture always controls the same semantic point like nose tip or canthus. In this way, facial geometry is independent of textures changes. We randomly shuffle face textures within a minibatch of the training set and render 3D faces back onto images as the inputs. Groundtruth 3DMM parameters and landmarks are intact. We show examples of the proposed face swapping augmentation in Fig. \ref{faceswap}. 

%For a sample in the training set, we shuffle the identity two times; that is, there are two swapped faces along with the original face associated with the same facial geometry.

% \begin{figure}[bt!]
%     \centering
%     \includegraphics[width=1.0\linewidth]{figure/face_swap_v2.png}
%     \vspace{-12pt}
%     \caption{\textbf{Examples of the face swapping data augmentation.} }
%     \vspace{-15pt}
%     \label{faceswap}
% \end{figure}

\begin{table*}[tb!]
\begin{center}
  \caption{\textbf{Ablative for facial alignment.} The first table is for AFLW2000-3D using original groundtruth annotation. The second table is for the reannotated version. '-' means the module is not used and the corresponding loss terms are not introduced. The first row setting without all the introduced modules contains only a backbone structure for 3DMM parameter regression.}
  \label{FAL_ab}
  \footnotesize
  \begin{tabular}[c]
  {|p{2.0cm}<{\centering\arraybackslash}|
  p{3.4cm}<{\centering\arraybackslash}|
  p{2.2cm}<{\centering\arraybackslash}|
  p{1.0cm}<{\centering\arraybackslash}|
  p{1.0cm}<{\centering\arraybackslash}|
  p{1.0cm}<{\centering\arraybackslash}|
  p{1.0cm}<{\centering\arraybackslash}|
  p{0.7cm}<{\centering\arraybackslash}|}
  \hline
  
    AFLW2000-3D Original  & M$^2$ Feature Aggregation for Refinement & Landmark Geometry Support  & 0 to 30 & 30 to 60 & 60 to 90 & All \\
    \hline
     &  - & -  & 2.99 & 3.80 & 4.86 & 3.88\\ 
     &  \Checkmark & -  & 2.68 & 3.32 & 4.35 & 3.49\\ 
     &  - & \Checkmark  & 2.69 & 3.57 & 4.69 & 3.65\\ 
     &  \Checkmark & \Checkmark &  \textbf{2.66} & \textbf{3.30} & \textbf{4.27} & \textbf{3.41}\\
    \hline
    \hline
    AFLW2000-3D Reannotated  & M$^2$ Feature Aggregation for Refinement & Landmark Geometry Support & 0 to 30 & 30 to 60 & 60 to 90 & All \\
    \hline
     &  - & -  & 2.34 & 2.99 & 4.27 & 3.20\\ 
     &  \Checkmark & -  & 2.24 & 2.67 & 3.76 & 2.89\\ 
     &  - & \Checkmark & 2.23 & 2.69 & 3.90 & 2.94\\
     &  \Checkmark & \Checkmark &  \textbf{2.16} & \textbf{2.61} & \textbf{3.66} & \textbf{2.81}\\ 
    \hline
  \end{tabular}
  \vspace{-22pt}
\end{center}
\end{table*}

\textbf{Test sets for facial alignment.}
Facial alignment is widely evaluated on AFLW2000-3D \cite{zhu2016face}, which contains the first 2000 images of AFLW \cite{koestinger11a}, and it is annotated with a \textit{68-point} landmark definition. There are two groundtruth sets of AFLW2000-3D, original and reannotated by LS3D-W \cite{bulat2017far}). Reannotated version bears better groundtruth quality. We separately report performance for the two versions to fairly compare with related work on this standard dataset. We also follow \cite{guo2020towards} to evaluate on the full AFLW set, which contains 21K images using a \textit{21-point} landmark definition. The two datasets are used for showing evaluation on different number of facial landmarks. 

%However, from the results in \cite{guo2020towards}, there is a large performance gap between AFLW2000 and ALFW. We find that it is caused by using a simple 68 to 21 point mapping at inference time, which brings much ambiguity. To alleviate the issue, we manually annotate the 21-point definition on 3D Basel Face Model, while referring to the landmark definition of the AFLW dataset. By reannotation, we largely reduce the performance gap between AFLW2000 and AFLW.

\textbf{Test sets for 3D face modeling.}
We evaluate 3D face reconstruction on AFLW2000-3D\cite{zhu2016face}, MICC Florence\cite{florence2011}, 300VW \cite{shen2015first}, and Artistic-Faces \cite{yaniv2019face} for both quantitative and qualitative results. AFLW2000-3D contains 2000 fitted 3D face models. Florence contains high-resolution real face scans of 53 individuals. 300VW collects talks or interviews from the web, and Artistic-Faces gathers artistic style faces. 

\textbf{Test sets for face orientation estimation.}
Previous works \cite{zhu2016face, feng2018joint, guo2020towards, guo2020towards, bulat2017far} conduct evaluation focusing on the facial alignment and 3D face modeling. To fully evaluate 3D facial geometry, we introduce face orientation estimation for evaluation and comparison. We use AFLW2000-3D that contains face orientation in a wide range.

%We also compare with prior work focusing on face pose estimation without 3D face reconstruction. They directly regress 3 Euler angles for training and evaluation.

\subsection{Facial Alignment Evaluation}
\label{FAL_section}

\textbf{Metrics.} Normalized mean error (NME) in Eq.(\ref{NME_all}) for sparse facial landmarks with Euclidean distance is reported.
\begin{equation}
\vspace{-5pt}
     \text{NME} = \frac{1}{T}\sum^T_{t=1}\frac{\|u_t-v_t\|_2}{B},
\label{NME_all}
\vspace{-5pt}
\end{equation}
where $u_t$ and $v_t$ are predicted landmarks and groundtruth landmarks. Note that there is no need to further project $u_t$, since pose prediction is included in 3DMM that makes both $u_t$ and $v_t$  aligned with input images. $T$ is the number of samples, and $B$ is bounding box size, square root of box areas, as the normalization term for each detected face.

\textbf{Ablation study.} We show three different ablation studies in this section to fully examine the contribution of each part in M$^3$-LRN. Our aim is to validate the use of multi-modal/representation fusion and analyze how each modality/representation contributes to final performance using simple and widely-used network operations.

%Instead of using complex network operations, our work focuses on extensive studies using  multi-modal/representation ability to fully exploit facial landmarks.

\textbf{(1)} We first conduct ablation studies of the proposed method on AFLW2000-3D. M$^2$ Feature Aggregation for landmark refinement and Landmark Geometry Support are examined. Following \cite{zhu2019face, guo2020towards, feng2018joint}, we report NME under three yaw angle ranges. The results are shown in Table \ref{FAL_ab}. 

From the table, one could observe that both landmark refinement and landmark geometry support stages contribute to the final performance for better facial landmark estimation. M$^2$FA adopts the advantage of multi-modal and multi-representation to refine landmarks. LGS extracts the geometric information embedded in the 3D landmarks and helps 3DMM prediction with representation cycle. From the third and fourth rows of both tables, directly predicting landmark geometry from raw landmarks without refinement only obtains limited performance gains. In contrast, based on finer landmarks, LGS can further improve the results. This validates our M$^3$-LRN design that both M$^2$FA and LGS are required to attain the best performance.

%From later comparison with other works, we show that PGS breaks through the performance bottleneck that other works using only images to predict 3DMM parameters stuck at. 

\textbf{(2)} We then study using different modalities at M$^2$FA and different parameter regression targets at LGS. For the former, we experiment multi-modal fusion to construct MMPF by using only point feature, point+image feature, and all modalities in Fig. \ref{mmfv} (point, image, and 3DMM modalities). For the latter, we examine performance of different regression targets at LGS for pose, shape, and expression. Results are shown in Table \ref{FAL_PIFV_ab}. Row 1-3 show that the performance gain of using image and 3DMM modalities mainly comes from small or medium pose ranges. This is because images and 3DMM from images capture more descriptive features on frontal face cases. Large pose cases would cause \textit{self-occlusion} on images and make prediction unreliable. Row 4-6 exhibit effects of regressing only pose (Row 4), shape and expression (Row 5), and all (Row 6). The improvements mainly come from larger pose cases. The reason is that LGS regresses 3DMM parameters from 3D landmarks that provide more expressive features on larger pose cases due to the \textit{representation advantage} that 3D points do not suffer from self-occlusion.

%where 2D representations suffer from heavy self-occlusion.  

\begin{table}[tb!]
\begin{center}
  \caption{\textbf{Study on different modalities and different regression targets for facial alignment.} AFLW2000-3D Original is adopted for evaluation. The first three rows study using different modalities to construct MMPF. Row 3 uses all modalities as in Fig. \ref{mmfv}. Based on the landmark refinement, Row 4-6 further study performance gains of different regression targets at the LGS stage.} \vspace{-5pt}
  \label{FAL_PIFV_ab}
  \footnotesize
  \begin{tabular}[c]
  {|
  p{3.08cm}<{\arraybackslash}|
  p{0.77cm}<{\centering\arraybackslash}|
  p{0.91cm}<{\centering\arraybackslash}|
  p{0.95cm}<{\centering\arraybackslash}|
  p{0.55cm}<{\centering\arraybackslash}|}
  \hline
      Structures  & 0 to 30 & 30 to 60 & 60 to 90 & All \\
    \hline
       Point feature only & 2.73 & 3.51 & 4.51 & 3.58\\ %[3.813, 4.420, 2.888, 3.707]
       Point + image feature & 2.68 & 3.40 & 4.61 & 3.56\\ 
       % [3.718, 4.394, 2.852, 3.655] 
       MMPF & 2.67 & 3.34 & 4.51 & 3.51\\ 
       %  [3.718, 4.368, 2.877, 3.654] 
       M$^2$FA+LGS (pose) & 2.68 & 3.31 & 4.55 & 3.51 \\
       M$^2$FA+LGS (shape, expr) & 2.67 & 3.32 & 4.47 & 3.48 \\
       M$^2$FA+LGS (all) &  \textbf{2.66} & \textbf{3.30} & \textbf{4.27} & \textbf{3.41}\\ 
       % [3.638, 4.070, 2.522, 3.410]
    \hline
  \end{tabular}
  \vspace{-13pt}
\end{center}
\end{table}

\begin{table}[tb!]
\begin{center}
  \caption{\textbf{Landmark geometry support network structure study.} Facial alignment on AFLW2000-3D Original is evaluated. Refer to Section \ref{FAL_section} for the setting details.}
  \label{PGS_study}
  \footnotesize
  \begin{tabular}[c]
  {|
  p{2.3cm}<{\centering\arraybackslash}|
  p{0.8cm}<{\centering\arraybackslash}|
  p{0.95cm}<{\centering\arraybackslash}|
  p{0.95cm}<{\centering\arraybackslash}|
  p{0.55cm}<{\centering\arraybackslash}|}
  \hline
      Settings & 0 to 30 & 30 to 60 & 60 to 90 & All \\
    \hline
       Setting 1 & 2.63 & 3.35 & 4.50 & 3.49\\ %[3.813, 4.420, 2.888, 3.707]
       Setting 2 & 2.62 & 3.31 & 4.31 & 3.41\\ 
       % [3.718, 4.394, 2.852, 3.655] 
       %Setting 3 & 2.63 & 3.32 & 4.31 & 3.42\\ 
       %  [3.718, 4.368, 2.877, 3.654] 
       LGS Setting & 2.66 & 3.30 & 4.27 & 3.41\\
       % [3.638, 4.070, 2.522, 3.410]
    \hline
  \end{tabular}
  \vspace{-15pt}
\end{center}
\end{table}

\textbf{(3)} We further investigate three different network designs of LGS. Setting 1 refines landmarks and regresses landmark geometry at the same step, and thus the landmark geometry is regressed from $L^c$ in this setting. This is to study whether LGS is benefited from refined landmarks $L^r$. Setting 2 further connects $z$, $\alpha_s$, and $\alpha_e$ to LGS in Fig. \ref{pipeline}, forming another multi-modal/representation feature aggregation stage to regress the landmark geometry. This is to analyze whether feature aggregation can also assist to predict landmark geometry. 

The study is shown in Table \ref{PGS_study}. Results of Setting 1 show that regressing landmark geometry from $L^c$ does not perform better than $L^r$ due to the finer and more accurate structure of $L^r$. We also find that multi-modal features used at LGS do not bring better performance in Setting 2. We assume this is because the information has been joined and utilized at the previous landmark refinement stage.

%Setting 3 is based on Setting 2. We further connect pose $P$ from the itemized 3DMM to PGS. Note that the pose information is not joined into MMFV since multi-modal point refinement targets at landmark refinement, and the input/output landmarks are at the same view  

%Pose is not added in Setting 2 since we focus on the landmark refinement, and the input and output at the refinement stage are at the same view. The third is to regress point geometry from landmark input, and we also add take information into consideration. 

%Most works report their numbers using the former, and fewer works evaluate on the latter.

\textbf{Comparison.} We exhibit benchmark comparison on popular and widely used AFLW2000-3D. As introduced earlier, the two versions of annotations (original and reannotated) are used. To have a fair comparison, we show results on the two annotations separately and compare with reported performances. Table \ref{FAL_AFLW2k_sota} shows the comparison on the original, and the reannotated version is in the supplementary. From these comparisons, our work attains the SOTA on this standard dataset. From the breakdown, our performance gain mainly comes from medium and large pose cases. We find that prior works encounter performance bottleneck, since they only utilize single modality/representation to regress 3DMM. However, referring to Table \ref{FAL_ab}, M$^2$FA has already shown the best performance compared with prior arts with its ability to fuse multi-modal and multi-representation features. Then the exploitation of LGS further shows lower error to further break through the performance bottleneck. % of previous works in Table \ref{FAL_AFLW2k_sota}.

%whose gain is not incremental when referring to the performance bottleneck in Table \ref{FAL_AFLW2k_sota}.

%we find that it is the exploitation of PGS to break through the performance bottleneck that other single modality and representation works encounter.

We show a larger performance gain on the reannotation version due to its better quality in the supplementary.

\begin{table}[tb!]
\begin{center}
  \caption{\textbf{Quantitative facial alignment comparison on benchmark.} AFLW2000-3D original annotation version is used. Our performance is the best with a gap over others on large poses.}
  \vspace{-7pt}
  \label{FAL_AFLW2k_sota}
  \footnotesize
  \begin{tabular}{|p{2.8cm}<{\centering}|
  p{0.85cm}<{\centering}|
  p{0.95cm}<{\centering}|
  p{0.95cm}<{\centering}|
  p{0.70cm}<{\centering}|}
  \hline
    AFLW2000-3D Original  &  0 to 30 & 30 to 60 & 60 to 90 & All \\
    \hline
    ESR \cite{cao2014face}& 4.60 & 6.70 & 12.67 & 7.99 \\
    %SDM \cite{xiong2015global}& 3.67 & 4.94 & 9.67 & 6.12 \\
    %3DDFA \cite{zhu2016face}& 3.78 & 4.54 & 7.93 & 5.42 \\
    3DDFA \cite{zhu2016face}& 3.43 & 4.24 & 7.17 & 4.94 \\
    Dense Corr \cite{yu2017learning}& 3.62 & 6.06 & 9.56 & 6.41 \\
    %DeFA \cite{liu2017dense} &  - & - & - & 4.50 \\
    3DSTN \cite{bhagavatula2017faster} & 3.15 & 4.33 & 5.98 & 4.49\\
    %CMD \cite{zhou2019dense} & - & -& - & 3.98 \\
    3D-FAN \cite{bulat2017far} & 3.16 & 3.53 & 4.60 & 3.76 \\
    3DDFA-PAMI \cite{zhu2019face} & 2.84 & 3.57 & 4.96 & 3.79 \\
    PRNet \cite{feng2018joint} & 2.75 & 3.51 & 4.61 & 3.62 \\
    2DASL \cite{tu20203d} & 2.75 & 3.46 & 4.45 & 3.55 \\
    3DDFA-V2 (MR)\cite{guo2020towards} & 2.75 & 3.49 & 4.53 & 3.59 \\
    3DDFA-V2 (MRS)\cite{guo2020towards} & \textbf{2.63} & 3.42 & 4.48 & 3.51 \\
    M$^3$-LRN (our) & \textbf{2.65} & \textbf{3.30} & \textbf{4.27} & \textbf{3.41} \\
    \hline
  \end{tabular}
  \vspace{-11pt}
  %\vspace{-27pt}
\end{center}
\end{table}

Following 3DDFA-V2\cite{guo2020towards}, we next perform evaluation on the AFLW full set (21K testing images with 21-point landmarks). We show the leaderboard comparison in Table \ref{FAL_AFLW}. Our work has the best performance, especially with a performance gap over others on large pose cases.

%As described in Section, we reannotate the corresponding points of 3D Basel Face model and 21 point definition to bridge the performance gap between AFLW2000 and AFLW. We compare with 3DDFA, PRNet, and 3DDFA-V2 and show the results in Table \ref{FAL_AFLW}. One could observe that the best-performing NME on AFLW 2000 reannotated version and AFLW are close, which shows a reasonable results, compared with results shown in \cite{guo2020towards}, since AFLW2000 is a regular subset of AFLW.

\begin{table}[tb!]
\begin{center}
  \caption{\textbf{Quantitative facial alignment comparison on AFLW with 21-point landmark definition.} }
  \label{FAL_AFLW}
  \footnotesize
  \begin{tabular}{|p{2.5cm}<{\centering}|
  p{0.85cm}<{\centering}|
  p{0.95cm}<{\centering}|
  p{0.95cm}<{\centering}|
  p{0.75cm}<{\centering}|}
  \hline
    AFLW  &  0 to 30 & 30 to 60 & 60 to 90 & All \\
    \hline
    ESR \cite{cao2014face} & 5.66 & 7.12 & 11.94 & 8.24 \\
    %SDM \cite{xiong2015global} & 4.75 & 5.55 & 9.34 & 6.55 \\
    3DDFA \cite{zhu2016face} & 4.75 & 4.83 & 6.39 & 5.32 \\
    3D-FAN \cite{bulat2017far} & 4.40 & 4.52 & 5.17 & 4.69 \\
    3DSTN \cite{bhagavatula2017faster} & \textbf{3.55} & \textbf{3.92} & 5.21 & 4.23 \\
    3DDFA-PAMI \cite{zhu2019face} & 4.11 & 4.38 & 5.16 & 4.55 \\
    PRNet \cite{feng2018joint} & 4.19 & 4.69 & 5.45 & 4.77\\
    3DDFA-V2 \cite{guo2020towards} & 3.98 & 4.31 & 4.99 & 4.43 \\
    M$^3$-LRN (our) & 3.76 & \textbf{3.92} & \textbf{4.48} & \textbf{4.06}\\
    \hline
  \end{tabular}
  \vspace{-11pt}
  %\vspace{-25pt}
\end{center}
\end{table}

\subsection{Face Orientation Estimation Evaluation}

\textbf{Metrics and studies.} Following the evaluation protocol in \cite{yang2019fsa, ruiz2018fine}, we calculate mean absolute error (MAE) of predicted Euler angles in degree. AFLW2000-3D with groundtruth angles for each face are used, except 31 samples whose yaw angles are outside the range [-99, 99]. We first study different combinations of the proposed method in Table \ref{FOE_ab} and report MAE of each angle separately. From the results, M$^2$FA contributes to finer landmark structures that result in better orientation estimation. LGS adopts pose parameter regression from 3D points, and thus leads to more robust estimation than from 2D images. We further study information fusion of using different modalities at M$^2$FA and examine different parameter regression targets at LGS in Table \ref{FOE_LF_ab}. This analysis shows that more accurate orientation estimation is mainly benefited from pose regression at LGS. This breakdown also explains the performance gain about LGS in Table \ref{FOE_ab}.

\begin{table}[tb!]
\begin{center}
  \caption{\textbf{Ablative for face orientation estimation.} Same modules are also studied in Table \ref{FAL_ab} for facial alignment. 
  MAE of Euler angles in degree is reported.}
  %\vspace{-7pt}
  \label{FOE_ab}
  \footnotesize
  \begin{tabular}[c]
  {|p{1.85cm}<{\centering\arraybackslash}|
  p{1.2cm}<{\centering\arraybackslash}|
  p{0.56cm}<{\centering\arraybackslash}|
  p{0.56cm}<{\centering\arraybackslash}|
  p{0.56cm}<{\centering\arraybackslash}|
  p{0.56cm}<{\centering\arraybackslash}|}
  \hline
  M$^2$ Feature Aggregation for Refinement & Landmark Geometry Support & Yaw & Pitch & Roll & Mean \\
    \hline
      - & - & 3.97 & 4.93 & 3.28 & 4.06\\ 
      \Checkmark & - & 3.72 & 4.37 & 2.88 & 3.65\\ 
      - & \Checkmark & 3.67 & 4.48 & 2.95 & 3.70\\ 
      \Checkmark & \Checkmark & \textbf{3.42} & \textbf{4.09} & \textbf{2.55} & \textbf{3.35}\\
    \hline
  \end{tabular}
  \vspace{-13pt}
  %\vspace{-22pt}
\end{center}
\end{table}

\begin{table}[tb!]
\begin{center}
  \caption{\textbf{Study on different modalities and different regression targets for face orientation estimation.} Same structures are also studied in Table \ref{FAL_PIFV_ab} for facial alignment. The first three rows study using different modalities to construct MMPF. Based on the landmark refinement, Row 4-6 further study performance gains of different regression targets at the LGS stage.}
  \vspace{-5pt}
  \footnotesize
  \label{FOE_LF_ab}
  \begin{tabular}[c]
  {|
  p{3.8cm}<{\arraybackslash}|
  p{0.65cm}<{\centering\arraybackslash}|
  p{0.65cm}<{\centering\arraybackslash}|
  p{0.65cm}<{\centering\arraybackslash}|
  p{0.6cm}<{\centering\arraybackslash}|}
  \hline
      Structures  & Yaw & Pitch & Roll & Mean \\
    \hline
       Point-feature only & 3.81 & 4.42 & 2.89 & 3.71\\
       Point + image feature & 3.72 & 4.39 & 2.85 & 3.66\\ 
       MMPF & 3.72 & 4.37 & 2.88 & 3.65\\
       M$^2$FA+LGS (pose) &  3.58 & 4.06 & 2.57 & 3.40\\ 
       M$^2$FA+LGS (shape, expr) & 3.47 & 4.23 & 2.59 & 3.43\\ 
       M$^2$FA+LGS (all) &  3.42 & 4.09 & 2.55 & 3.35\\ 
    \hline
  \end{tabular}
  \vspace{-15pt}
\end{center}
\end{table}

\textbf{Comparison.} We next show benchmark comparison. We collect works that focus only on face orientation estimation \cite{ruiz2018fine, yang2019fsa} and 3DMM-based methods \cite{zhu2016face, zhu2019face, guo2020towards, tu20203d} for 3D face modeling. The 3DMM-based works do not include evaluation of this task. In order to show the full benchmark list, we evaluate their methods using their pretrained models. Table \ref{FOE_sota} shows that our method is the best with a nearly 20\% significant improvement. Overall 3D-based methods are more robust than pure 2D methods. We display visual comparison in Fig. \ref{compare_FAL_FOE} and more studies, discussion, and results in the supplementary.  

\begin{figure}[tb!]
    \centering
    \vspace{-3pt}
    \includegraphics[width=0.85\linewidth]{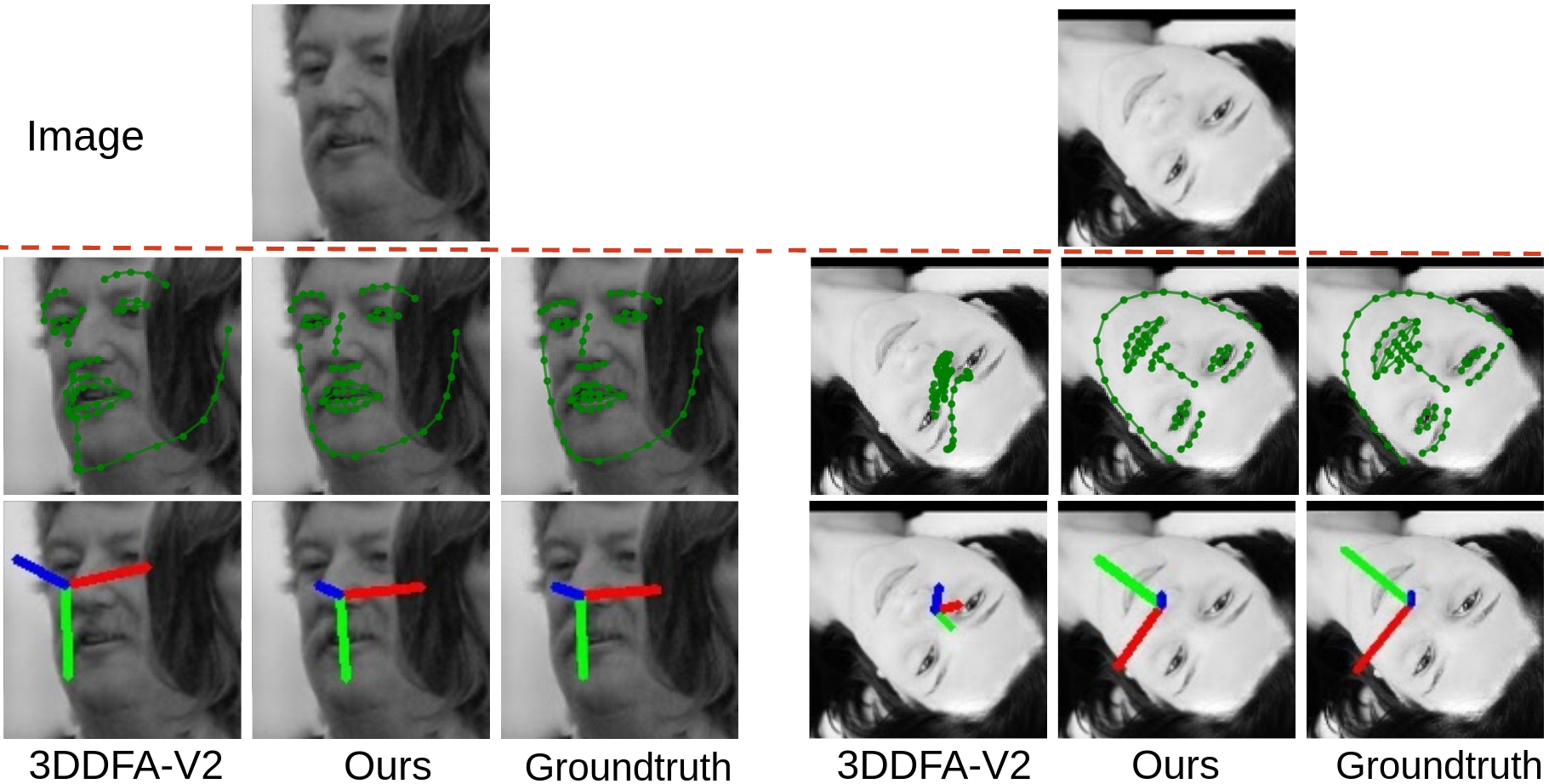}
    
    \caption{\textbf{Qualitative comparison of facial alignment and orientation estimation.} Note that unlike 2D landmarks, 3D landmarks can predict occluded face outlines. The left case is low-resolution and blurry and thus challenging. The right case is with rare and extreme roll rotation. Our results show more robustness over 3DDFA-V2.}
    \label{compare_FAL_FOE}
    \vspace{-5pt}
\end{figure}

\begin{table}[tb!]
\begin{center}
  \caption{\textbf{Face orientation estimation benchmark comparison on ALFW2000-3D.} PnP shows solving perspective n-point problems using groundtruth landmarks. Note that we do not include PRNet here because it does not infer face orientation directly and also obtains poses by PnP with predicted landmarks.}
  \vspace{0pt}
  \label{FOE_sota}
  \footnotesize
  \begin{tabular}{|p{3.0cm}<{\centering}|
  p{0.8cm}<{\centering}|
  p{0.8cm}<{\centering}|
  p{0.8cm}<{\centering}|
  p{0.75cm}<{\centering}|}
  \hline
    AFLW2000-3D  & Yaw & Pitch & Roll & Mean \\
    \hline
    PnP-landmark & 5.92 & 11.76 & 8.27 & 8.65 \\
    FAN-12 point \cite{bulat2017far}& 6.36 & 12.30 & 8.71 & 9.12 \\
    HopeNet \cite{ruiz2018fine}& 6.47 & 6.56 & 5.44 & 6.16 \\
    SSRNet-MD \cite{ssrnet}& 5.14 & 7.09 & 5.89 & 6.01 \\
    FSANet \cite{yang2019fsa}& 4.50 & 6.08 & 4.64 & 5.07 \\
    3DDFA-TPAMI \cite{zhu2019face} & 4.33 & 5.98 & 4.30 & 4.87 \\
    2DASL \cite{tu20203d} & 3.85 & 5.06 & 3.50 & 4.13\\
    3DDFA-V2 \cite{guo2020towards} & 4.06 & 5.26 & 3.48 & 4.27 \\
    M$^3$-LRN (our) & \textbf{3.42} & \textbf{4.09} & \textbf{2.55} & \textbf{3.35} \\
    \hline
  \end{tabular}
  %\vspace{-15pt}
\end{center}
\end{table}

\subsection{3D Face Modeling Evaluation}
\label{3dface_recon}
\begin{figure}[tb!]
    \centering
    \vspace{-5pt}
    \includegraphics[width=0.85\linewidth]{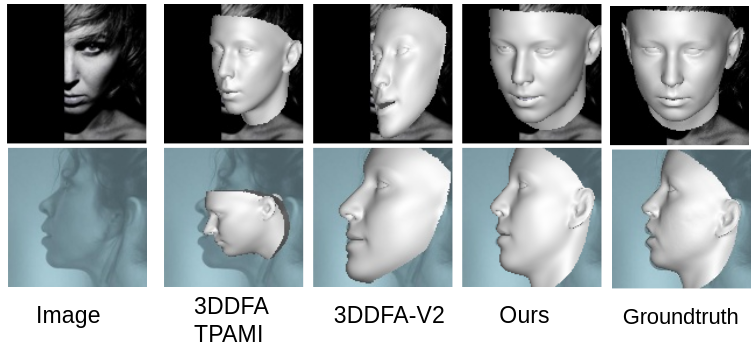}
    \caption{\textbf{Qualitative comparison of 3D face models.} Our results are robust to cropped face and underwater face examples.}
    \label{Qual_3DFace}
    \vspace{-14pt}
\end{figure}
\textbf{Metrics and Comparison}. Following \cite{feng2018joint, tu20203d, guo2020towards}, we first evaluate 3D face modeling on AFLW2000-3D. Two protocols are used. Protocol 1 uses the iterative closet point (ICP) algorithm to register groundtruth 3D models and predicted models. NME of per-point error normalized by interocular distances is calculated \cite{feng2018joint, tu20203d, zhu2016face}. Protocol 2, also called dense alignment, calculates the per-point error normalized by bounding box sizes with groundtruth models aligned with images \cite{guo2020towards}. Since ICP is not used, pose estimation would affect the performance under this protocol, and the NME would be higher. We illustrate numerical comparison in Table \ref{ALFW2K_Recon}. The results show our ability of M$^3$-LRN to recover 3D face models from monocular inputs and attain the best performance. In addition, we further exhibit visual comparison in Fig. \ref{Qual_3DFace}. Our M$^3$-LRN is capable of recovering 3D faces under extreme scenarios, such as heavily cropped or underwater cases. In contrast, other works produce skewed faces. 

\begin{table}[tb!]
\begin{center}
  \caption{\textbf{3D face modeling comparison on AFLW2000-3D.} Two protocols are applied. Refer to Section \ref{3dface_recon} for the details.}
  %\vspace{-11pt}
  \label{ALFW2K_Recon}
  \footnotesize
  \begin{tabular}{|p{1.3cm}<{\centering}|
  p{0.8cm}<{\centering}|
  p{0.8cm}<{\centering}|
  p{0.7cm}<{\centering}|
  p{0.8cm}<{\centering}|
  p{1.1cm}<{\centering}|}
    \hline
     Protocol-1 \cite{feng2018joint, tu20203d, zhu2016face}& 3DDFA\cite{zhu2016face} & DeFA\cite{liu2017dense} & PRNet\cite{feng2018joint} & 2DASL\cite{tu20203d} & M$^3$-LRN (our) \\
    \cline{1-6}
    NME & 5.37 & 5.55 & 3.96 & 2.10 & \textbf{1.97}\\
    \hline
  \end{tabular}
  \begin{tabular}{|p{1.5cm}<{\centering}|
  p{0.8cm}<{\centering}|
  p{0.8cm}<{\centering}|
  p{1.6cm}<{\centering}|
  p{1.3cm}<{\centering}|}
    \hline
    Protocol-2 \hspace{1cm}\cite{guo2020towards} & 3DDFA \cite{zhu2016face} & DeFA\cite{liu2017dense} & 3DDFA-V2 \hspace{0.5cm}\cite{guo2020towards} & M$^3$-LRN (our) \\
     \cline{1-5}
    NME & 6.56 & 6.04 & 4.18 & \textbf{4.06}\\
    \hline
  \end{tabular}
  \vspace{-13pt}
  %\vspace{-15pt}
\end{center}
\end{table}

Next, we evaluate the performance of 3D face modeling on Florence \cite{florence2011} with real scanned 3D faces. We follow the protocol from \cite{guo2020towards, feng2018joint}, which renders 3D face models on different views with pitch of -15, 20, and 25 degrees and yaw of -80, -40, 0, 40, and 80 degrees. The rendered images are used as the test inputs. After reconstruction, face models are cropped to 95mm from the nose tip, and ICP is performed to calculate point-to-plane root mean square error (RMSE) with cropped groundtruth. We show numerical results in Table \ref{Florence_sota}. We also display an error curve to show our robustness to yaw angle changes and a qualitative comparison for evaluation on Florence in the supplementary. %\vspace{-8pt}
\begin{table}[tb!]
\begin{center}
  \caption{\textbf{3D face modeling on Florence.} Point-to-plane RMSE is calculated for evaluation.}
  \vspace{-7pt}
  \footnotesize
  \label{Florence_sota}
  \begin{tabular}{|p{1.2cm}<{\centering}|
  p{0.8cm}<{\centering}|
  p{1.0cm}<{\centering}|
  p{1.7cm}<{\centering}|
  p{1.5cm}<{\centering}|}
  \hline
    Florence & PRNet\cite{feng2018joint} & 2DASL\cite{tu20203d} & 3DDFA-V2 \quad\quad \cite{guo2020towards} & M$^3$-LRN\quad(our) \\
    \hline
    RMSE & 2.25 & 2.05 & 2.79 & \textbf{1.87} \\
    \hline
  \end{tabular}
  \vspace{-20pt}
  %\vspace{-30pt}
\end{center}
\end{table}

\section{Conclusion}
\vspace{-4pt}
This work proposes a \textit{multi-task}, \textit{multi-modal}, and \textit{multi-representation} landmark refinement network to attain accurate 3D facial geometry prediction, including 3D facial alignment via 3D face models and face orientation estimation. The full exploitation of landmark information is key in our M$^3$-LRN. We adopt multi-modal/representation fusion for landmark refinement. Then, we are the first to study landmark geometry to help 3D facial geometry prediction. Further, the representation forms a cycle consistency.

Extensive experiments validate our network design. Specially, our M$^3$-LRN only adopts simple network operations with multi-modal/representation fusion to attain the SOTA. Our M$^3$-LRN is accurate, fast, and easy to implement. 

{\small
\bibliographystyle{ieee_fullname}
\bibliography{egbib}
}

%\end{document}
%%%%% Please comment out the following blocks to recover the review version. The following is the appended supp. 

\clearpage
\newpage
\pagebreak
\begin{center}
\textbf{\large Supplemental Materials:}
\end{center}
%%%%%%%%%% Merge with supplemental materials %%%%%%%%%%
%%%%%%%%%% Prefix a "S" to all equations, figures, tables and reset the counter %%%%%%%%%%
\setcounter{section}{0}
\setcounter{equation}{0}
\setcounter{figure}{0}
\setcounter{table}{0}
\setcounter{page}{1}
\makeatletter
\renewcommand{\theequation}{S\arabic{equation}}
\renewcommand{\thefigure}{S\arabic{figure}}
\renewcommand{\thetable}{S\arabic{figure}}
\renewcommand\thesection{\Alph{section}}
\renewcommand\thesubsection{\thesection.\Alph{subsection}}
%%%%%%%%%% Prefix a "S" to all equations, figures, tables and reset the counter %%%%%%%%%%

\section{Overview}

We document this supplementary in the following sections. In Section \ref{c}, we provide details of our network architectures and loss weights for training. We further present a study on backbone choices and a study to show our performance gain is not simply from using more parameters. In Section \ref{e},  evaluation of facial alignment on AFLW2000-3D reannotation version is exhibited. In Section \ref{g}, we present analysis and discussion on landmark geometry support. In Section \ref{f}, we conduct another experiment for face orientation estimation using the BIWI dataset. In Section \ref{h}, an error curve and visual comparison of 3D face modeling on Florence are illustrated. In Section \ref{b}, we describe texture synthesis using introduced UV-texture GAN and further compare with textures from 3DMM fitting. In Section \ref{i}, we add more qualitative results from our face geometry prediction using the 300VW video dataset and Artistic Faces.

\begin{figure*}[hbt]
    \centering
    \includegraphics[width=1.0\linewidth]{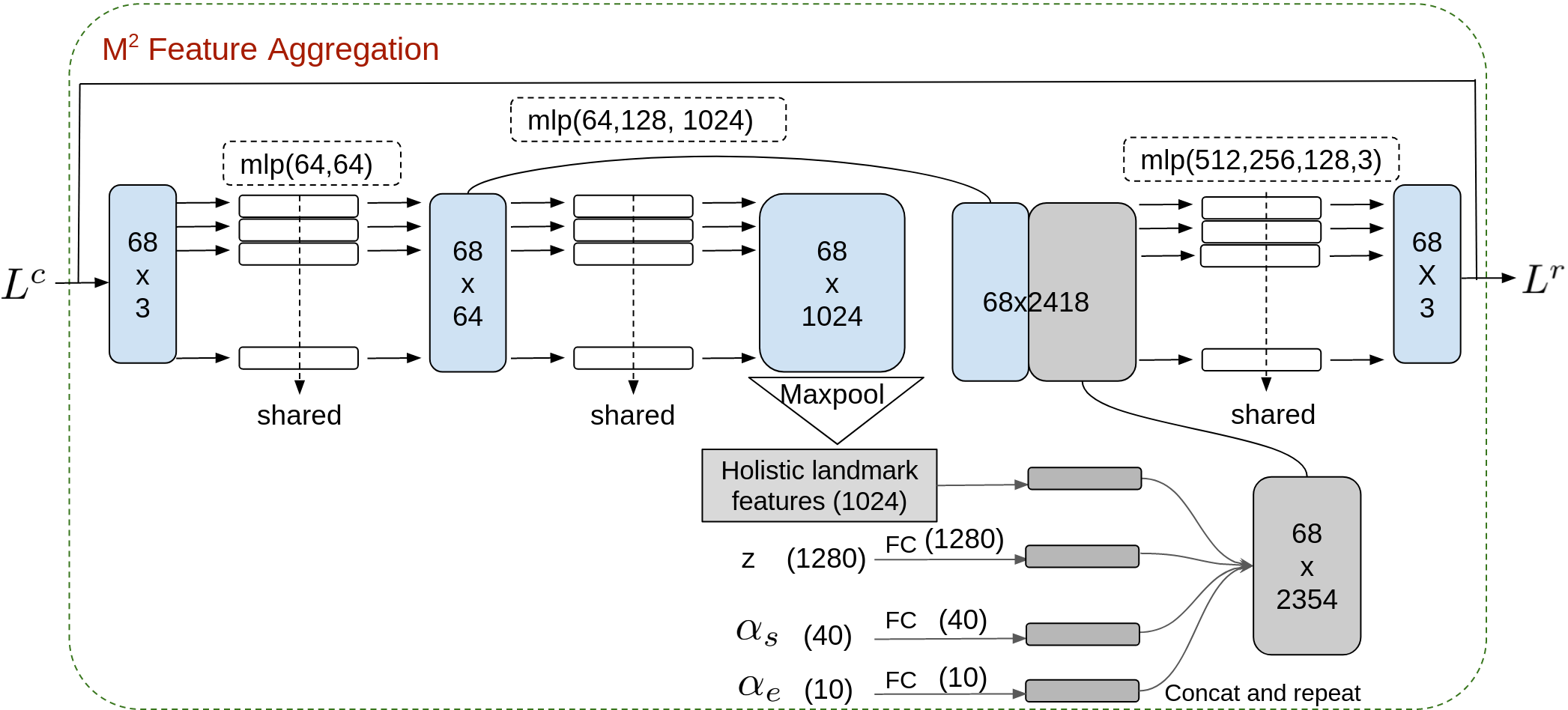}
    \caption{\textbf{Detailed Structure of M$^2$FA.} mlp(64,64) means two MLP layers with output channel sizes 64 and 64. ReLU and batch normalization are used for each layer. The notations correspond to those in the main paper. ($z$ is latent image feature, $\alpha_s$ and $\alpha_e$ are shape and expression 3DMM parameters from the first stage, $L^c$ and $L^r$ are 3D landmarks before and after the landmark refinement. }
    \label{m2fa}
    \vspace{-5pt}
\end{figure*}

\begin{figure}[hbt]
    \centering
    \includegraphics[width=1.0\linewidth]{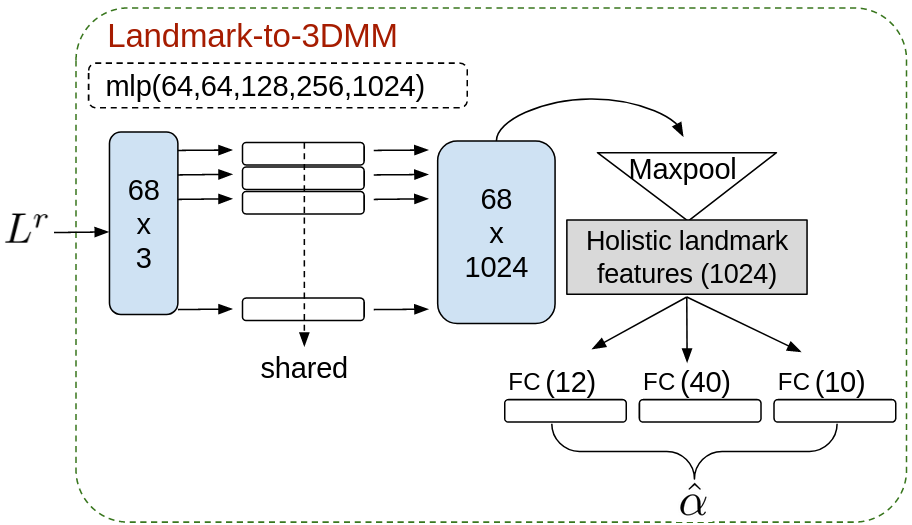}
    \caption{\textbf{Detailed Structure of the landmark-to-3DMM module.} The notations correspond to those in the main paper. $L^r$ is 3D landmarks after the refinement and $\hat{\alpha}$ is the regressed landmark geometry.}
    \label{lgs}
    \vspace{-5pt}
\end{figure}

\section{Network Architecture, Hyper-Parameters, and Network Parameter Studies}
\label{c}

\textbf{Details of architecture.} Based on Fig. 2 in the main paper (pipeline graph of our M$^3$-LRN), our network architecture details are described as follows. We use MobileNet-V2 as the face encoder backbone following \cite{guo2020towards}. The latent image features $z$ after the global max-pooling is 1280-dim. The pose, shape, and expression decoders are fully-connected (FC) layers with input $z$ and output 3DMM parameters of 12 ($\alpha_p$), 40 ($\alpha_s$), and 10 ($\alpha_e$) dimensions for pose, shape, and expression.

We next illustrate the detailed structures of M$^2$FA in Fig. \ref{m2fa}.
Aggregation of the latent image features, $\alpha_s$, and $\alpha_e$ and global point features form a 2354-dim feature vector. We repeat and append this vector to the low-level point features to obtain MMPF of 68$\times$2418. Later we use another MLPs to obtain refined landmarks $L^r$ from MMPF. 

We then illustrate the detailed structure of the landmark-to-3DMM module in Fig. \ref{lgs}. Given $L^r$ from the landmark refinement stage, this module regresses 3DMM parameters $\hat{\alpha}$.

\textbf{Loss weights.} Beside hyperparameters described in the main paper, for weights of the loss combination, we choose $\lambda_1 = 0.02$, $\lambda_2 = 0.03$, $\lambda_3 = 0.02$, and $\lambda_4 = 0.001$ for training.

\textbf{Study on the backbone.} We next conduct a study on the backbone network choices using the AFLW full set. We select MobileNet-V2 \cite{sandler2018mobilenetv2}, ResNet50 \cite{he2016deep}, ResNet101 \cite{he2016deep}, ResNeSt50 \cite{zhang2020resnest}, and  ResNeSt101\cite{zhang2020resnest} for comparison. From Table \ref{FAL_AFLW_backbone}, one could see that the 101-residual layer network is too deep, and thus the performance drops compared with the 50-residual layer network. ResNeSt, a split-attention variant of ResNet, can improve the performance for the 101-residual layer case, but the performance of ResNeSt50 is on par with ResNet50. We think this is because the split-attention scales better and remedies the undesirable effects of deeper networks, which are described in their work \cite{zhang2020resnest}.

\begin{table}[tb!]
\begin{center}
  \caption{\textbf{Network backbone study on the AFLW full set.} We compare MobileNet \cite{sandler2018mobilenetv2}, ResNet \cite{he2016deep}, and ResNeSt \cite{zhang2020resnest}, a ResNet variant with split-attention.}
  \vspace{-10pt}
  \label{FAL_AFLW_backbone}
  \begin{tabular}{|p{2.0cm}<{\centering}|
  p{1.0cm}<{\centering}|
  p{1.2cm}<{\centering}|
  p{1.2cm}<{\centering}|
  p{0.75cm}<{\centering}|}
  \hline
    Backbone  &  0 to 30 & 30 to 60 & 60 to 90 & All \\
    \hline
    MobileNet & 3.86 & 4.13 & 4.61 & 4.20 \\
    ResNet50 & 3.76 & 3.92 & 4.48 & 4.06 \\
    ResNeSt50 & 3.76 & 3.92 & 4.52 & 4.07 \\
    ResNet101 & 3.90 & 4.14 & 5.08 & 4.38\\
    ResNeSt101 & 3.78 & 4.04 & 4.62 & 4.15\\
    \hline
  \end{tabular}
  \vspace{-15pt}
\end{center}
\end{table}

\textbf{Study on the network parameters.} We further conduct a study to verify the effectiveness of the proposed M$^2$FA and landmark-to-3DMM modules. In the following experiment, our aim is to show that the performance gain comes from the design of the proposed modules in our framework, rather than simply using more network parameters.

By using the MobileNet-V2 backbone, network parameters of our M$^3$-LRN amount to 3.8M (3.0M for the first-stage 3DMM regression, and 0.8M for the M$^2$FA and landmark-to-3DMM modules). We build another basic model consists of only the first-stage 3DMM regressor with more network parameters. We add MLPs with ReLU and BN that amount to 0.8M parameters after the latent image code $z$ to regress $\alpha_p$, $\alpha_s$, and $\alpha_e$. Therefore, this basic model and our M$^3$-LRN have approximately the same number of network parameters. Then we experiment on facial alignment and face orientation estimation for the basic model and M$^3$-LRN.

From the results in Table \ref{network_params}, more parameter adoption in the first-stage 3DMM regressor only leads to minor improvements. Especially for facial alignment, using extra 0.8M parameters of MLPs only gives 0.02 overall performance gain. The results validate our proposed M$^2$FA and landmark-to-3DMM designs, showing that the performance gain of our M$^3$-LRN is not from simply using more parameters. 

\begin{table*}[htb!]
\begin{center}
  \caption{\textbf{Comparison with the first-stage 3DMM regressor with extra parameters.} The first table shows results on AFLW2000-3D Original for facial alignment, and the second table shows results also on AFLW2000-3D for face orientation estimation. \# of params means the number of network parameters. More parameter use in the first-stage 3DMM regressor only results in limited performance gain. The experiment validates designs of the proposed M$^2$FA and landmark-to-3DMM modules in our M$^3$-LRN. }
  %\vspace{-10pt}
  \label{network_params}
  \begin{tabular}{|p{6.5cm}<{\centering}|
  p{2.0cm}<{\centering}|
  p{1.2cm}<{\centering}|
  p{1.2cm}<{\centering}|
  p{1.2cm}<{\centering}|
  p{0.75cm}<{\centering}|}
  \hline
    Structures  & \# of params & 0 to 30 & 30 to 60 & 60 to 90 & All \\
    \hline
    Only first-stage regressor & 3.0M & 2.99 & 3.80 & 4.86 & 3.88 \\
    Only first-stage regressor (w/ extra MLPs) & 3.8M & 2.98 & 3.82 & 4.77 & 3.86 \\
    M$^3$-LRN & 3.8M & 2.66 & 3.30 & 4.27 & \textbf{3.41} \\
    \hline
  \hline
      Structures & \# of params & Yaw & Pitch & Roll & Mean \\
    \hline
       Only first-stage regressor & 3.0M & 3.97 & 4.93 & 3.28 & 4.06\\
       Only first-stage regressor (w/ extra MLPs) & 3.8M & 3.80 & 4.62 & 2.84 & 3.75\\ 
       M$^3$-LRN & 3.8M & 3.42 & 4.09 & 2.55 & \textbf{3.35}\\
    \hline
  \end{tabular}
  \vspace{-15pt}
\end{center}
\end{table*}

\section{Evaluation on Reannotated AFLW2000-3D}
\label{e}
Reannotation of AFLW2000-3D is provided in LS3D-W\cite{bulat2017far}. Few works, DHM \cite{sun2018deep} and MCG-Net \cite{shang2020self}, report their performance on the annotated version. To aggregate more results for this study, we also include evaluation of 3DDFA \cite{zhu2016face}, PRNet \cite{feng2018joint}, and 3DDFA-V2 \cite{guo2020towards} using their pretrained models. From Table \ref{FAL_reanno}, normalized mean errors (NMEs) are generally lower than using the original annotation. This shows the higher quality of the reannotation. Among the methods for comparison, our result holds the best performance and has a clear performance gap over others. Compared with PRNet, the second-best method in the table, our improvements are derived from large pose cases. %This shows that we successfully utilize the multi-modal and multi-representation advantage for to attain better alignment performance.

\begin{table}[tb!]
\begin{center}
  \caption{\textbf{Comparison on AFLW2000-3D Reannotation.} Our method has the best alignment result and holds a clear performance gap over others.}
  \label{FAL_reanno}
  \vspace{-5pt}
  \begin{tabular}{|p{2.25cm}<{\centering}|
  p{1.0cm}<{\centering}|
  p{1.2cm}<{\centering}|
  p{1.2cm}<{\centering}|
  p{0.65cm}<{\centering}|}
  \hline
    AFLW2000-3D Reannotated  &  0 to 30 & 30 to 60 & 60 to 90 & All \\
    \hline
    DHM \cite{sun2018deep} & 2.28 & 3.10 & 6.95 & 4.11 \\
    {3DDFA \cite{zhu2016face}} & 2.84 & 3.52 & 5.15 & 3.83 \\
    PRNet \cite{feng2018joint} & 2.35 & 2.78 & 4.22 & 3.11 \\
    MGCNet \cite{shang2020self} & 2.72 & 3.12 & 3.76 & 3.20 \\
    3DDFA-V2 \cite{guo2020towards} & 2.84 & 3.03 & 4.13 & 3.33 \\
    M$^3$-LRN (our) & \textbf{2.16} & \textbf{2.60} & \textbf{3.64} & \textbf{2.80} \\
    \hline
  \end{tabular}
  \vspace{-18pt}
\end{center}
\end{table}

\section{Discussion on Landmark Geometry Support}
\label{g}
\textbf{Why using sparse landmarks rather than full vertices?} Landmark geometry $\hat{\alpha}$ in Sec.3.3 of the main paper describes embedded facial geometric information in sparse 3D landmarks. In contrast to sparse landmarks (68 points in our work), mesh from BFM Face includes 53.5K vertices (45K if excluding the neck and ears). When surveying on point processing, much research adopts only 1024 or 2048 points \cite{qi2017pointnet, qi2017pointnet++, xu2020grid}. Much denser points are inefficient for point processing, and the accommodation is also limited by GPU memory. On the other hand, because facial alignment is considered as an upstream task for the downstream such as face recognition \cite{shi2006effective} or recent streaming video compression \cite{NvidiaMaxine, wang2020one}, high efficiency is more desirable.

Although a point-sampling strategy could be used for downsizing, 3D landmarks are very efficient and compact for expressing facial traits and outlines. Therefore, 3D landmarks are desirable for predicting facial geometry, and our focus of this work is to exert the speciality of landmarks by a multi-modal/representation feature extraction and fusion framework.

%The information is considered auxiliary since compared with dense face mesh, sparse 3D landmarks provide approximate facial geometry.

\textbf{How sparse landmarks assist 3DMM regression?} Compared with 2D images, 3D sparse landmarks describe facial traits and approximate face outlines. Although landmarks are sparse, the representation provides another view to learn facial geometry and complements with 3DMM regressed from images. For example, facial geometry for large pose cases is hard to estimate from the 2D due to self-occlusion. Further, the face orientation is defined in the 3D space; thus, it is more advantageous to estimate face orientation from 3D points and attains less ambiguous learning paradigm. From Table 2 and 7 in the main paper that both study the contributions for each regression target at LGS, MMPF+LGS(all) improves the performance compared with using MMPF alone on both evaluation of facial alignment and face orientation estimation. \textit{The results show the ability of regressing landmark geometry $\hat{\alpha}$ and use $\hat{\alpha}$ to self-supervise $\alpha$ from the first-stage 3DMM regression.} 

We also evaluate the performance of using $\hat{\alpha}$ as the output on facial alignment and face orientation estimation using AFLW2000-3D. The 3D landmarks reconstructed by $\hat{\alpha}$ and the face orientation converted from its pose parameter $\hat{\alpha_p}$ attain an NME of 6.48 on the alignment and an MAE of 5.76 on the orientation estimation. The results are reasonable, since the direct input source to the landmark-to-3DMM module is only 68-point sparse landmarks, which presents only approximate facial traits and outlines. However, these numerical results are still comparable with some methods in Table 4 and 8 of the main paper. This validates our training strategy that \textit{3DMM estimation from sparse landmarks as the direct input achieve on par performance with some studies using regression from images}, which contain more complete face information.

%\textit{using the 3DMM parameters from only 68-point sparse landmarks as the direct input, the performance is comparable to some previous studies whose regressions are from dense images}.

%since $\hat{\alpha}$ is regressed from only 68-point sparse landmarks showing only approximate facial geometry.

\section{Face Orientation Estimation on BIWI}
\label{f}

We further use BIWI head pose dataset \cite{fanelli_IJCV} for the evaluation. Following \cite{yang2019fsa}, the BIWI testing set contains 8 individuals with 8 separate sequences from Kinect and about 5K testing images in total. Each individual is asked to rotate their heads to left, right, up, and down. FaceBoxes \cite{zhang2017faceboxes} is used to detect and crop out faces. We compare with works that trains on 300W-LP and evaluates on BIWI Test, including HopeNet \cite{ruiz2018fine}, SSR-Net-MD\cite{ssrnet}, FSA-Net\cite{yang2019fsa}. From Table \ref{biwi_comp}, our performance is the best on the leaderboard \cite{yang2019fsa} based on the MAE of Euler angles. 

\begin{table}[tb!]
\begin{center}
  \caption{\textbf{Comparison on BIWI head pose dataset for face orientation estimation.} $\alpha$ here is a weight in HopeNet. Our M$^3$-LRN holds the best performance. }
  \vspace{-5pt}
  \label{biwi_comp}
  \begin{tabular}[c]
  {|
  p{3.2cm}<{\centering\arraybackslash}|
  p{0.8cm}<{\centering\arraybackslash}|
  p{0.8cm}<{\centering\arraybackslash}|
  p{0.8cm}<{\centering\arraybackslash}|
  p{0.8cm}<{\centering\arraybackslash}|}
  \hline
      Method  & Yaw & Pitch & Roll & Mean \\
    \hline
       HopeNet\cite{ruiz2018fine} ($\alpha =1$) & 4.81 & 6.61 & 3.27 & 4.90\\
       HopeNet\cite{ruiz2018fine} ($\alpha =2$) & 5.17 & 6.98 & 3.39 & 5.18\\ 
       SSR-Net-MD\cite{ssrnet} & 4.49 & 6.31 & 3.61 & 4.65\\
       FSA-Caps-Fusion\cite{yang2019fsa} &  4.27 & 4.96 & \textbf{2.76} & 4.00\\ 
       M$^3$-LRN (our) & \textbf{3.99} & \textbf{4.78} & 2.93 & \textbf{3.90}\\
    \hline
  \end{tabular}
  \vspace{-15pt}
\end{center}
\end{table}

\section{3D Faces on Florence}
\label{h}
\begin{figure}[bt!]
    \centering
    \includegraphics[width=1.0\linewidth]{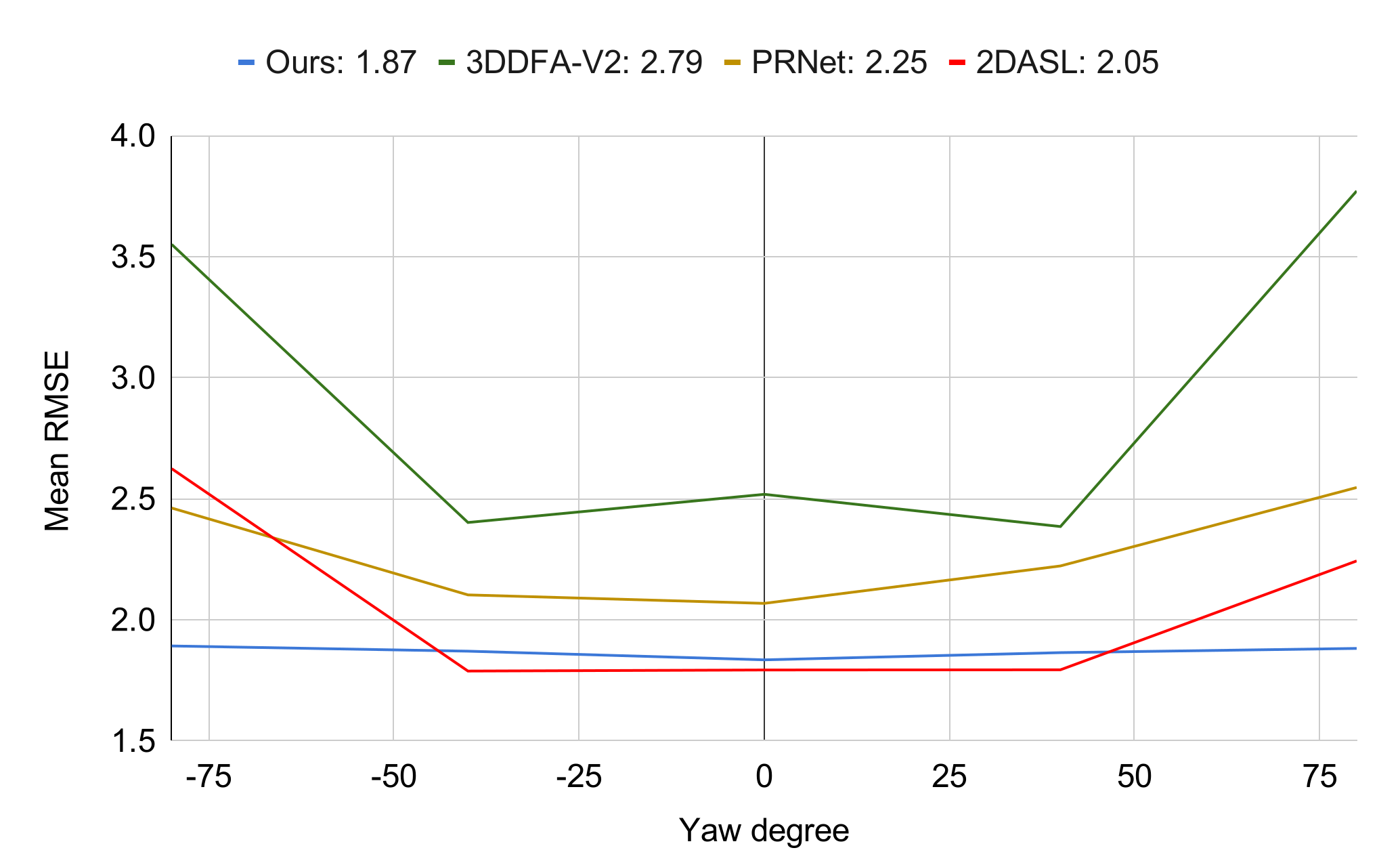}
    \caption{\textbf{Error curve for 3D face modeling by yaw angle on the Florence dataset.} Our method is rather robust to pose changes and attains the lowest mean point-to-plane RMSE. }
    \label{florence}
    \vspace{-1pt}
\end{figure}

Based on the Florence experiment in Section 4.3 and Table 10 of the main paper, we further show an error curve comparison in Fig. \ref{florence} for 3D face modeling. Our method is rather robust to pose changes and attains a \textit{nearly but not totally flat} error curve. Although the results of 2DASL for low and medium cases are close to ours, they are not robust for large pose cases.

We visualize the reconstructed meshes and compare with the current best-performing 3DMM-based method (3DDFA-V2 \cite{guo2020towards}) and UV-position-based method (PRNet \cite{feng2018joint}) in Fig. \ref{florence_recon_comp}. We mark point-to-plane RMSEs beside each face model results. From the upper example, our reconstructed faces show narrower eye-to-side distances, higher cheeks, and pointed chins. These features are consistent with the groundtruth model. On the other hand, 3DDFA-V2 shows wider faces, unapparent cheeks, and non-pointed chins; thus, their errors are higher. Besides, for a cropping range of 95mm from the nose tip, 3DDFA-V2 shows more forehead areas than the groundtruth model, which means the geometry prediction is not accurate. PRNet is not 3DMM-based. Although PRNet has higher flexibility to predict per-vertex deformation due to its non-parametric nature, it is also harder to estimate a precise 3D face via vertex regression on a UV-position map. From the lower example, our faces are wider and consistent with the groundtruth. In contrast, 3DDFA-V2 shows more elongated shapes for large pose cases. PRNet shows skewed faces under large pose scenarios.

\section{Texture Synthesis}
\label{b}
\begin{figure}[bt!]
    \centering
    \includegraphics[width=1.0\linewidth]{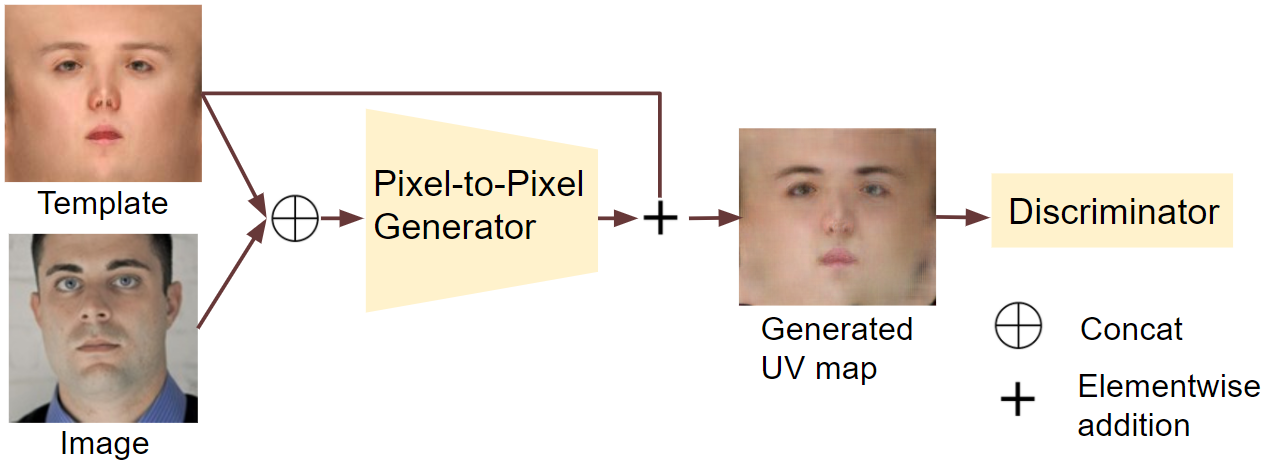}
    \caption{\textbf{UV-texture GAN.} The generator produces a UV map from a fixed template and an input image. The generated UV map combines structures of the template and skin color of the image.}
    \label{uvgan}
    \vspace{-3pt}
\end{figure}

Most previous works for 3D facial alignment via 3D face modeling mainly focus on the geometry \cite{zhu2016face, zhu2019face, guo2020towards, feng2018joint, tu20203d}. To get more realistic 3D face models, here we also conduct a smaller study on texture synthesis that lies upon our predicted 3D face models.

\begin{figure}[bt!]
    \centering
    \includegraphics[width=1.0\linewidth]{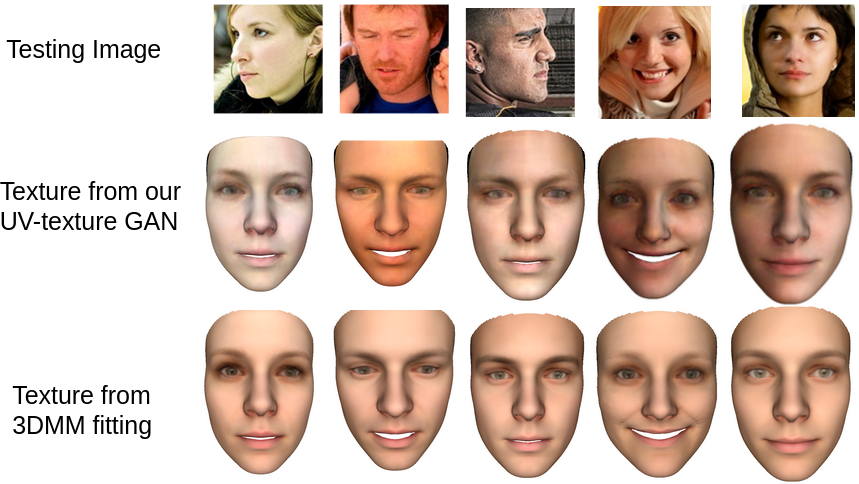}
    \caption{\textbf{Synthesized texture comparison.} Texture synthesized by the introduced UV-texture GAN is more realistic than textures from 3DMM fitting.}
    \label{uv_compare}
    \vspace{-5pt}
\end{figure}

Similar to 3DMM fitting for 3D faces, as illustrated in the main paper Eq.(1), textures can also be synthesized by adding a mean texture term and a multiplication term of texture basis and parameters. For example, the 300W-LP and AFLW2000-3D datasets contain texture parameters with a 199-dim texture basis. However, some limitations exist due to its 3DMM-ish style. First, variations and diversities of generated facial textures directly rely on the texture basis's expressiveness. Precise calculation from large and diverse datasets is required to attain high quality. Next, 3DMM is a parameterization technique, generated textures lie on manifolds spanned by the low-rank basis matrix. Therefore, 3DMM texture fitting usually produces over-smooth textures that lack reality. A visualization is illustrated in Fig. \ref{uv_compare}. 
%Other works on 3D facial alignment focus on facial geometry and pay less attention to texture synthesis for their 3D models \cite{zhu2016face, zhu2019face, guo2020towards, feng2018joint, tu20203d}.

Here we introduce a simple but effective UV-texture Generative Adversarial Network (UV-texture GAN) for texture synthesis. The model structure is illustrated in Fig. \ref{uvgan}. UV mapping \cite{foley1996computer} involves per-vertex color mappings from UV-texture maps. Each vertex is associated with its $(u,v)$-coordinate for querying vertex color from the three-channel UV-texture maps.

\begin{figure}[bt!]
    \centering
    \includegraphics[width=0.90\linewidth]{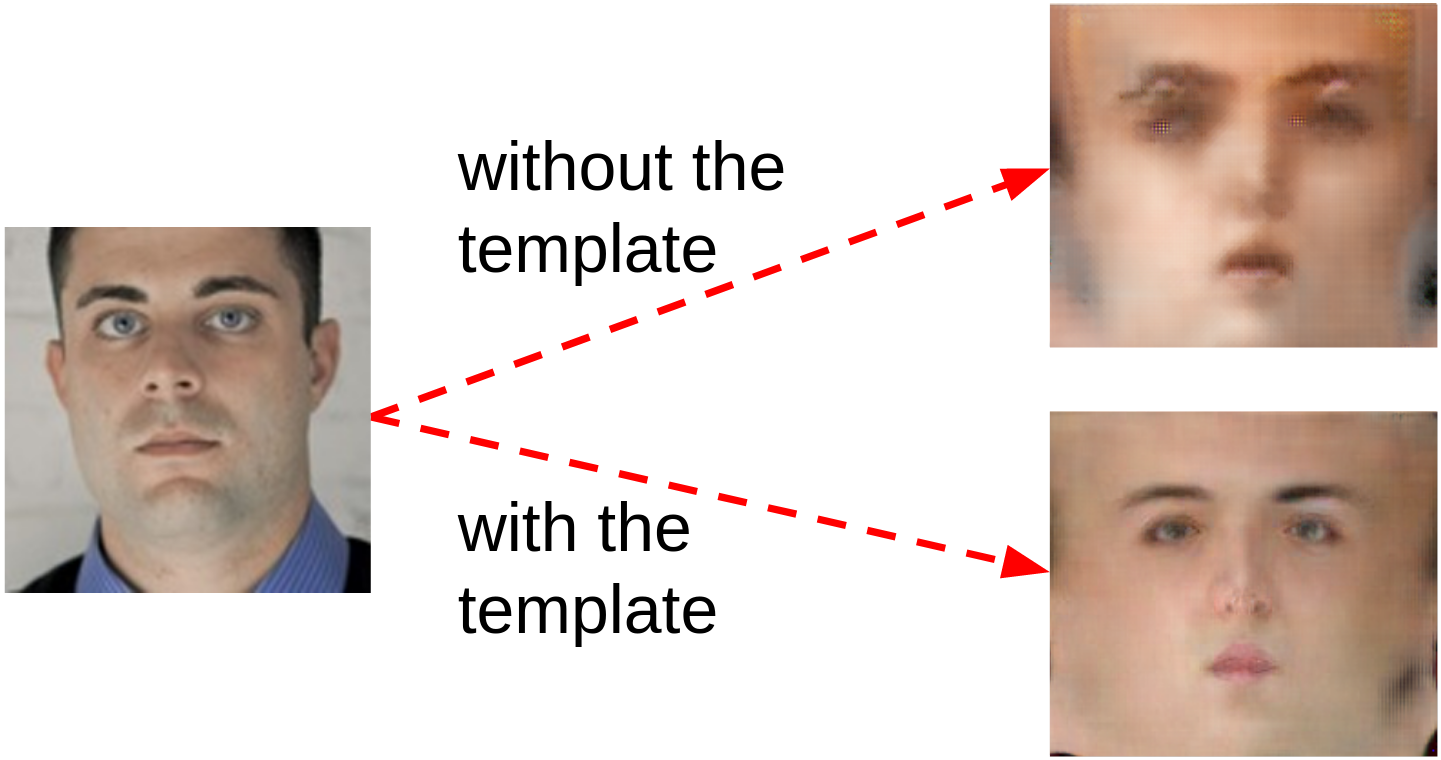}
    \caption{\textbf{Effects of using the template in the UV-texture GAN}. Without the template shown in Fig. \ref{uvgan}, facial traits such as eyes and moth are blurry and their positions are inaccurate.}
    \label{template_compare}
    \vspace{-4pt}
\end{figure}

\begin{figure*}[bt!]
    \centering
    \includegraphics[width=0.85\linewidth]{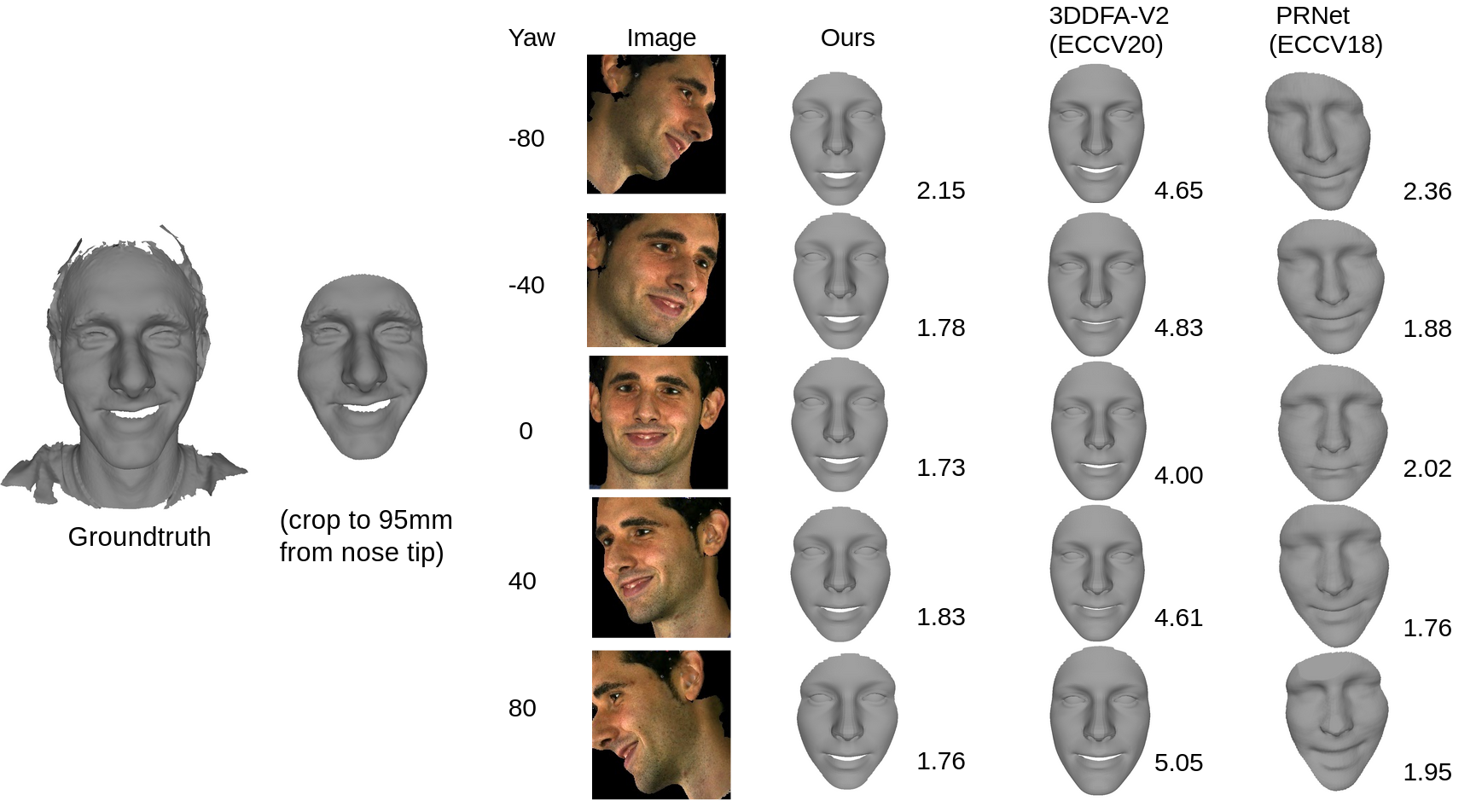}
    \includegraphics[width=0.85\linewidth]{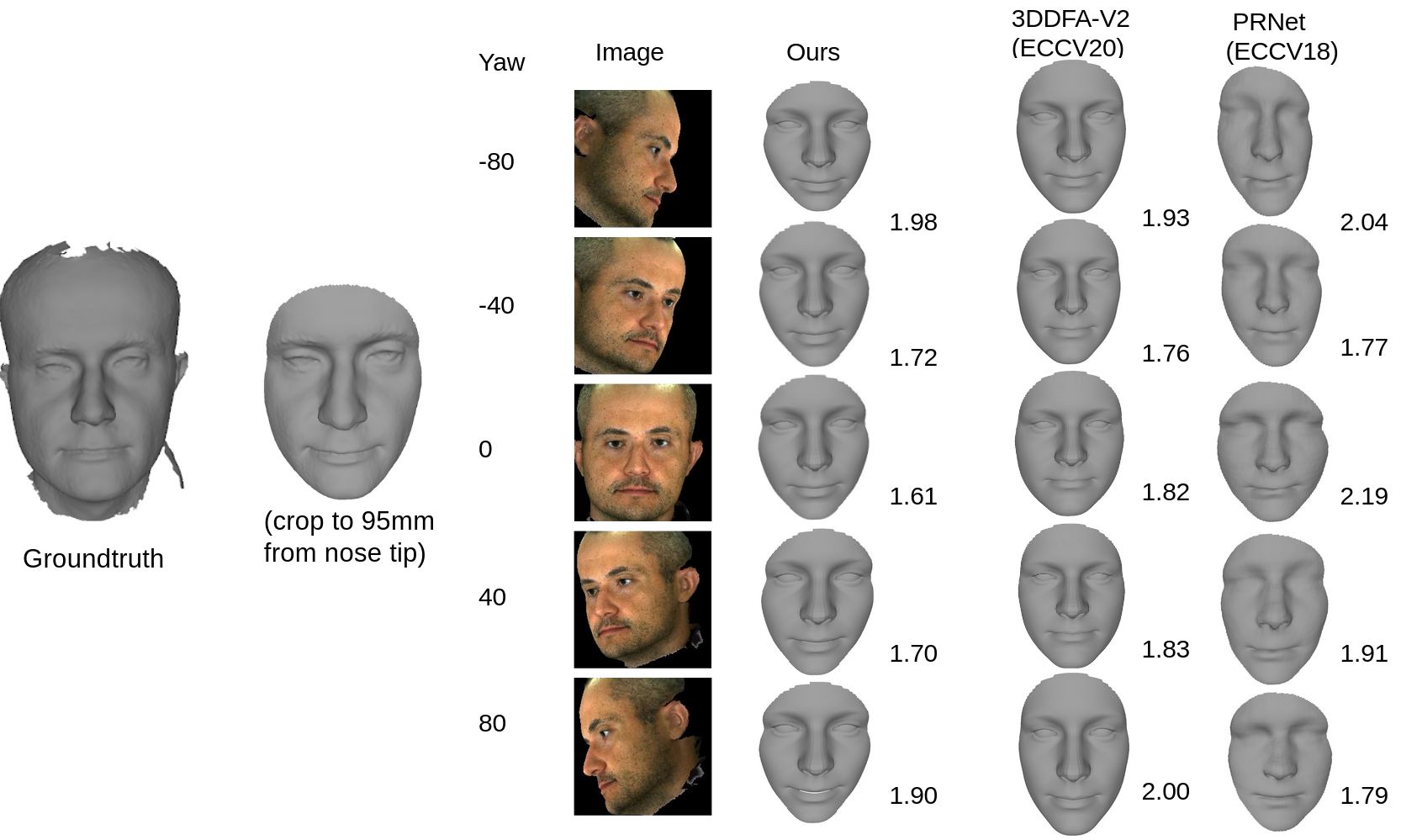}
    \caption{\textbf{Reconstructed face comparison by yaw angle on examples from the Florence dataset.} Number beside the reconstructed models are their normalized point-to-plane RMSEs. Our results are robust to pose changes. For the upper example, 3DDFA-V2 shows wider faces, unapparent cheeks, non-pointed chins, and larger forehead areas with a cropping range of 95mm from the nose tip; therefore, their results hold higher errors. PRNet shows imprecise facial structures. In addition, their faces are twisted for large pose cases. For the lower example, groundtruth face is wider, but 3DDFA-V2 shows more elongated faces, and PRNet predictions are not reliable. Our results are more similar to the groundturh shape.}
    \label{florence_recon_comp}
\end{figure*}

The introduced UV-texture GAN adopts a pixel-to-pixel image translator that transforms unstructured in-the-wild images to a canonical UV space to generate the UV maps from images. However, pixel-to-pixel style transfer \cite{isola2017image, wang2018pix2pixHD} retains input image structures, such as salient object outlines, and produce a different style artifact. It is hard to map unconstrained face images onto the canonical UV space by direct pixel-to-pixel translators. To resolve the issue, we further feed a template UV map together with a face image as inputs to the generator (Fig. \ref{uvgan}). The template is projected from the mean texture of BFM Face. Further, we shortcut the template to the output for facilitating the training procedure, where the generator learns a mapping from the six-channel input to the residual UV space. We display the ability of the template in Fig. \ref{template_compare}.

To form our training set, we collect about 2000 in-the-wild frontal face images and warp the faces onto the UV space with the aids of facial landmarks. ResNet is used as the generator backbone. Least square GAN (LSGAN) is used for the loss. We train the network with 300 epochs. Adam is adopted as the optimizer with an initial learning rate of 0.0002, which linearly drops to 0 after 100 epochs.

We show comparison in Fig. \ref{uv_compare} to illustrate the difference of synthesized textures between using the introduced UV-texture GAN and 3DMM texture parameter estimation. Results from the UV-texture GAN are more realistic and not over-smooth that, drawing a contrast with results from 3DMM fitting. Skin colors are more similar to images, since the introduced UV-texture GAN combines hue from images and structures from the template to produces more realistic textures. 

\section{More Qualitative Results}
\label{i}
Here we further show more qualitative results from the 300VW dataset for talks or interview videos \cite{shen2015first} in Fig. \ref{300vw-1}, \ref{300vw-2}, \ref{300vw-3}, \ref{300vw-4} and Artistic Faces (AF) for different artistic style faces \cite{yaniv2019face} in Fig. \ref{af-1}, \ref{af-2}.

\begin{figure*}[bt!]
    \centering
    \includegraphics[width=1.0\linewidth]{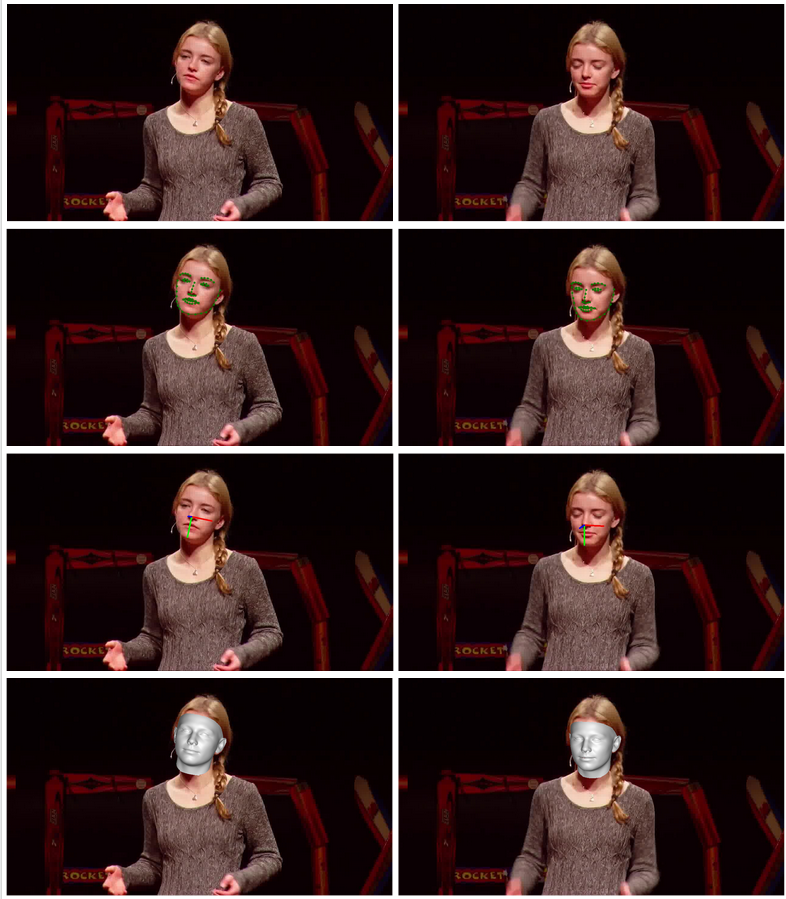}
    \caption{\textbf{Results of 3D geometry prediction on 300VW from our method.} Row 1-4: images, landmarks, face orientation, 3D faces.}
    \label{300vw-1}
\end{figure*}

\begin{figure*}[bt!]
    \centering
    \includegraphics[width=1.0\linewidth]{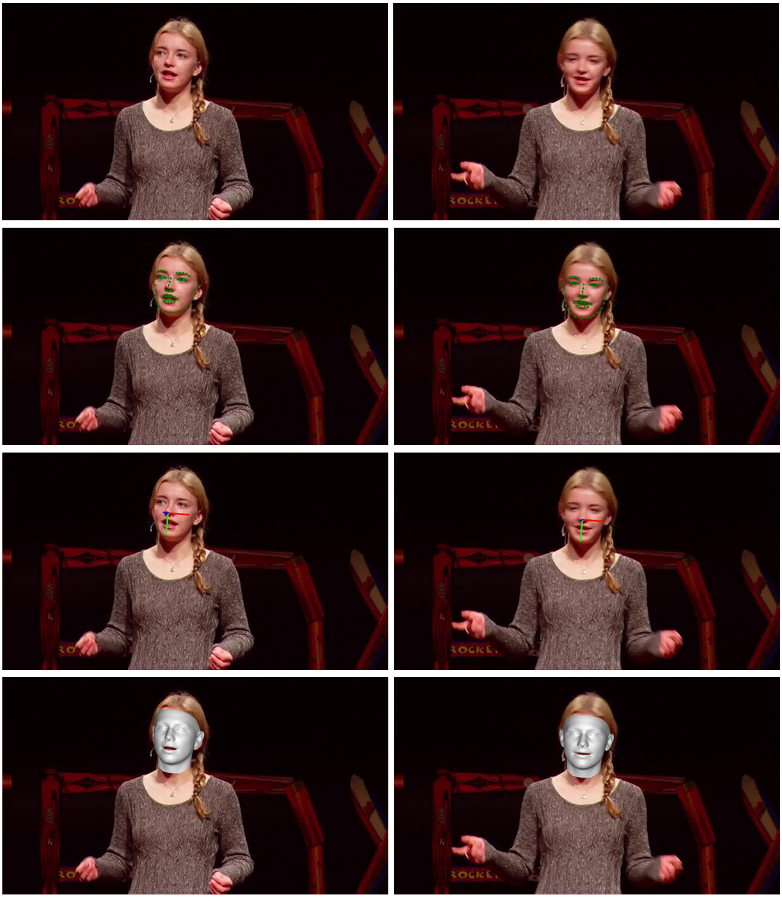}
    \caption{\textbf{(Continued) Results of 3D geometry prediction on 300VW from our method.} Our result is robust to motion blur for the right-hand-side case.}
    \label{300vw-2}
\end{figure*}

\begin{figure*}[bt!]
    \centering
    \includegraphics[width=1.0\linewidth]{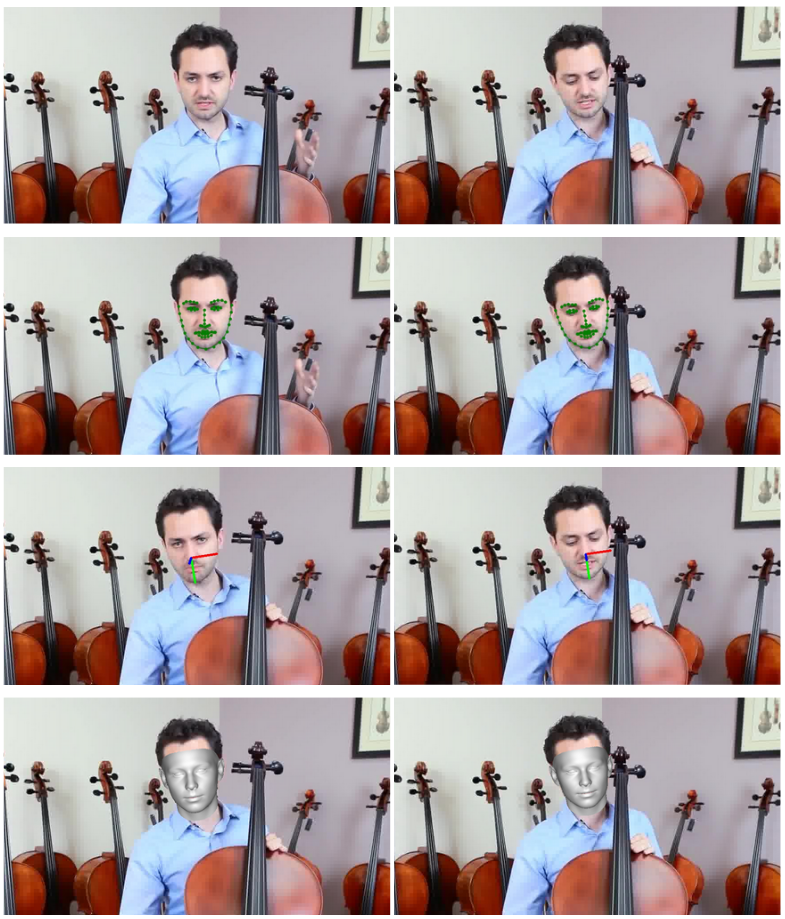}
    \caption{\textbf{(Continued) Results of 3D geometry prediction on 300VW from our method.}}
    \label{300vw-3}
\end{figure*}

\begin{figure*}[bt!]
    \centering
    \includegraphics[width=1.0\linewidth]{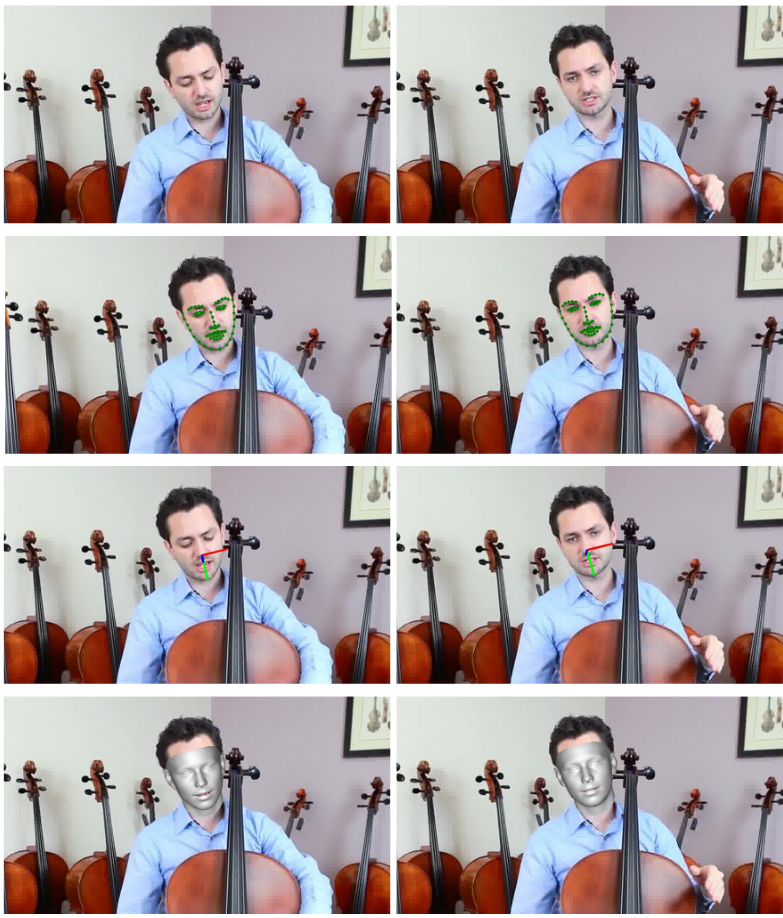}
    \caption{\textbf{(Continued) Results of 3D geometry prediction on 300VW from our method.}}
    \label{300vw-4}
\end{figure*}

\begin{figure*}[bt!]
    \centering
    \includegraphics[width=1.0\linewidth]{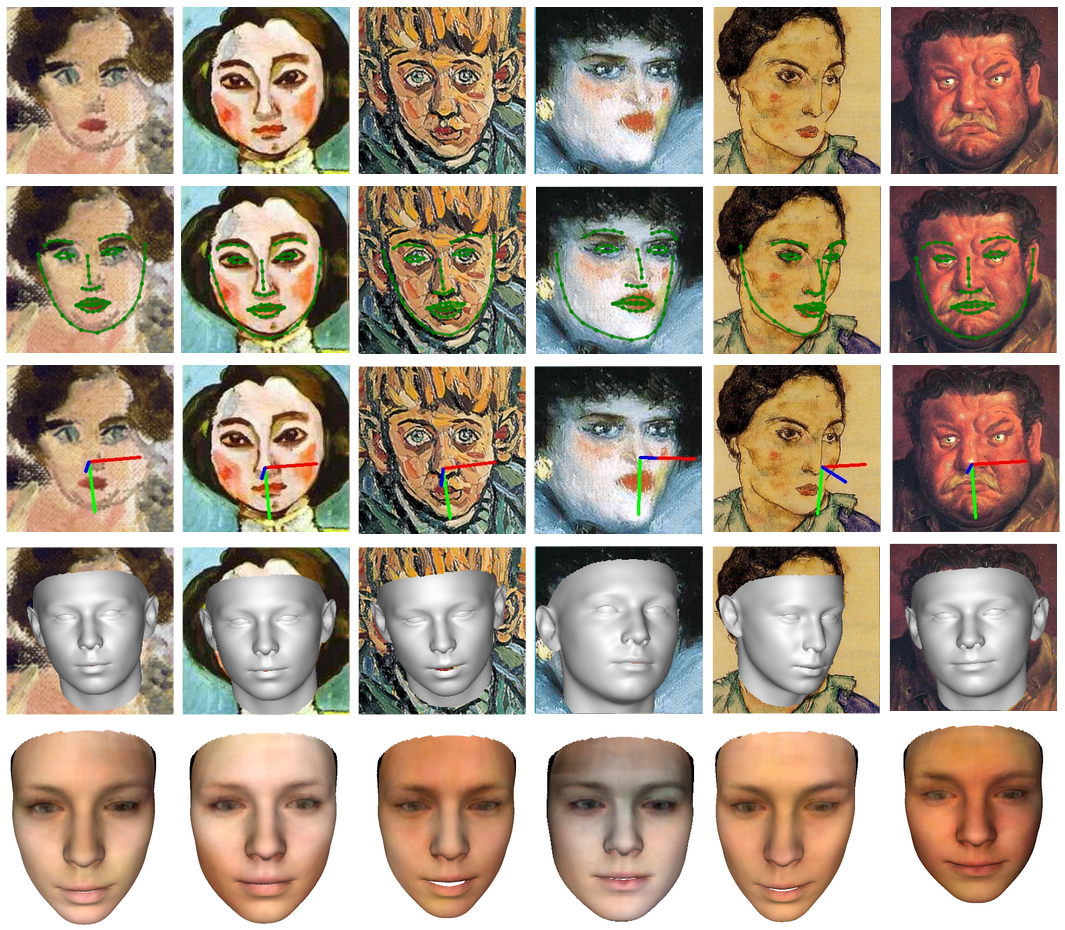}
    \caption{\textbf{Results of 3D geometry prediction on Artistic Faces from our method.} Row 1-5: images, landmarks, face orientation, 3D faces, textures.}
    \label{af-1}
\end{figure*}

\begin{figure*}[bt!]
    \centering
    \includegraphics[width=1.0\linewidth]{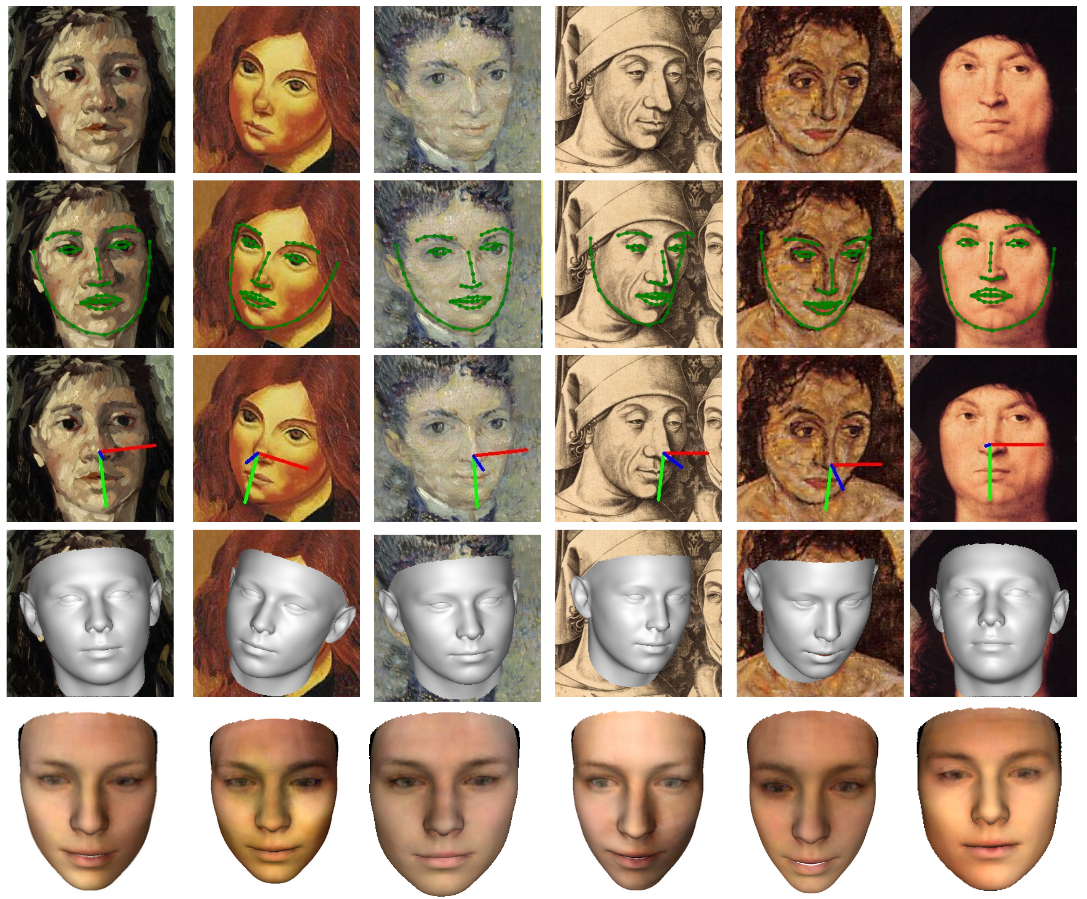}
    \caption{\textbf{(Continued) Results of 3D geometry prediction on Artistic Faces from our method.}}
    \label{af-2}
\end{figure*}

% \clearpage
% \clearpage
% \clearpage
% {\small
% \bibliographystyle{ieee_fullname}
% \bibliography{egbib_supp}
% }

\end{document}